%% file: main.tex
\documentclass{article} 
\usepackage{iclr,times}

\usepackage[utf8]{inputenc} 
\usepackage[T1]{fontenc}    
\usepackage{hyperref}       
\usepackage{url}            
\usepackage{booktabs}       
\usepackage{amsfonts}       
\usepackage{nicefrac}       
\usepackage{microtype}      
\usepackage{xcolor}         
\usepackage[english]{babel}
\usepackage{multirow}
\usepackage{graphicx}
\usepackage{adjustbox}
\usepackage{enumitem}
\usepackage{algorithm}
\usepackage[noend]{algpseudocode}
\usepackage{algorithmicx}
\usepackage{subcaption}
\usepackage{soul}

\usepackage{caption}
\definecolor{hypercolor}{HTML}{1f72d1}
\hypersetup{
    colorlinks=true,
    linkcolor=hypercolor,
    urlcolor=hypercolor,
    citecolor=hypercolor,
    filecolor=hypercolor,
}
\setlist{leftmargin=0.25in}

\input{math_macros.tex}

\title{A Genetic Algorithm for Navigating \\ Synthesizable Molecular Spaces}

\author{Alston Lo, Connor W. Coley, Wojciech Matusik \\
Massachusetts Institute of Technology \\
\texttt{\{alston,wojciech\}@csail.mit.edu}, \; \texttt{ccoley@mit.edu} \\
}

\newcommand{\ours}{SynGA}
\newcommand{\oursbo}{SynGBO}

\iclrfinalcopy 
\AtBeginDocument{\lhead{Published as a conference paper at ICLR 2026}}

\begin{document}

\maketitle

\begin{abstract}
Inspired by the effectiveness of genetic algorithms and the importance of synthesizability in molecular design, we present SynGA, a simple genetic algorithm that operates directly over synthesis routes. Our method features custom crossover and mutation operators that explicitly constrain it to synthesizable molecular space. By modifying the fitness function, we demonstrate the effectiveness of SynGA on a variety of design tasks, including synthesizable analog search and sample-efficient property optimization, for both 2D and 3D objectives. Furthermore, by coupling SynGA with a machine learning-based filter that focuses the building block set, we boost SynGA to state-of-the-art performance. For property optimization, this manifests as a model-based variant SynGBO, which employs SynGA and block filtering in the inner loop of Bayesian optimization. Since SynGA is lightweight and enforces synthesizability by construction, our hope is that SynGA can not only serve as a strong standalone baseline but also as a versatile module that can be incorporated into larger synthesis-aware workflows in the future. Our code is available at \url{https://github.com/alstonlo/synga}.
\end{abstract}

\input{sections/0_introduction}
\input{sections/1_background}
\input{sections/2_approach}
\input{sections/3_experiments}
\input{sections/4_discussion}

\bibliography{references}
\bibliographystyle{main}

\newpage
\appendix
\input{sections/appendix}

\end{document}

%% file: math_macros.tex
\usepackage{amsmath}
\usepackage{amssymb}
\usepackage{amsthm}
\usepackage{mathtools}

\usepackage{pifont}
\newcommand{\cmark}{\ding{51}}
\newcommand{\xmark}{\ding{55}}

\newcommand{\synspace}{\Mcal_S}
\newcommand{\minus}{\scalebox{0.75}[1.0]{$-$}}

\newcommand{\pmocell}[1]{{\footnotesize #1}}

\newcommand{\N}{\mathbb{N}}

\newcommand{\R}{\mathbb{R}}

\newcommand{\Bcal}{\mathcal{B}}

\newcommand{\Dcal}{\mathcal{D}}

\newcommand{\Fcal}{\mathcal{F}}

\newcommand{\Hcal}{\mathcal{H}}

\newcommand{\Lcal}{\mathcal{L}}
\newcommand{\Mcal}{\mathcal{M}}
\newcommand{\Ncal}{\mathcal{N}}
\newcommand{\Ocal}{\mathcal{O}}
\newcommand{\Pcal}{\mathcal{P}}

\newcommand{\Rcal}{\mathcal{R}}
\newcommand{\Scal}{\mathcal{S}}
\newcommand{\Tcal}{\mathcal{T}}

\renewcommand{\vec}[1]{\mathbf{#1}}

\newcommand{\vb}{\vec{b}}

\newcommand{\vq}{\vec{q}}

\newcommand{\vx}{\vec{x}}
\newcommand{\vy}{\vec{y}}

\DeclarePairedDelimiterX{\innerp}[2]{\langle}{\rangle}{#1, #2}

\DeclareMathOperator{\arity}{arity}

%% file: sections/0_introduction.tex
\section{Introduction}

The design of novel molecules is a costly and time-intensive endeavor, so significant effort has gone into developing computational tools to de-risk and accelerate the process. Molecular design involves a constrained optimization problem that is made challenging by the discrete combinatorial nature of molecular space and the need for sample-efficiency. The rapid development of machine learning (ML) has led to exciting advances for in silico design, with methods such as variational autoencoders \citep{bombarelli2018automatic}, reinforcement learning \citep{reinvent}, GFlowNets \citep{bengio2021flow}, and large language models \citep{chemcrow} being proposed, to name a few. Yet among them, genetic algorithms (GAs) \citep{holland1992adaptation}, a classical approach, have remained competitive for their simplicity, sample-efficiency, and exploratory power \citep{molga,pmo}. This is in contrast to standard ML methods which tend to be data-hungry and struggle to extrapolate from their training sets. GAs have been used to design organic emitters \citep{nigam2024emitters}, polymers \citep{kim2021polymer}, catalysts \citep{seumer2024beyond}, and drugs \citep{terfloth2001neural}. However, unlike ML models, classical GAs cannot learn insights from data and are reliant on expert-designed genetic operators. Thus, there is increasing interest in enhancing GAs with ML (or vice versa) \citep{kneilding2024augmenting}. This is primarily done by (1) using the GA as a subroutine within some broader ML workflow, or (2) augmenting a part of the GA (e.g., crossover) with ML (Section \ref{sec:background}). Such work  reaffirms the strength of GAs in chemistry and shows that GAs and ML can in fact be coupled synergistically.  

Search power, however, matters only if domain constraints are obeyed.  Many molecular generative models are synthesis-agnostic, which can lead to them proposing unstable or unsynthesizable designs \citep{gao2020synthesizability}. This presents a major barrier for adopting these models in real-world applications, regardless of their performance on benchmarks. While using retrosynthesis models post-hoc can alleviate this issue, they may also incur prohibitive runtime, taking minutes per evaluation \citep{aizynthfinder}. Instead, another promising strategy is to incorporate synthesis considerations directly into the model itself \citep{stanley2023fake}. One class of synthesis-aware models are those that operate directly on synthesis routes, often defined in terms of a fixed catalog of purchasable building blocks and expert-defined reaction templates. These template-based models are appealing because they are explicitly constrained and the molecules produced by them come automatically with plausible synthesis routes.

\begin{figure}[t]
\centering
\includegraphics[width=0.98\textwidth]{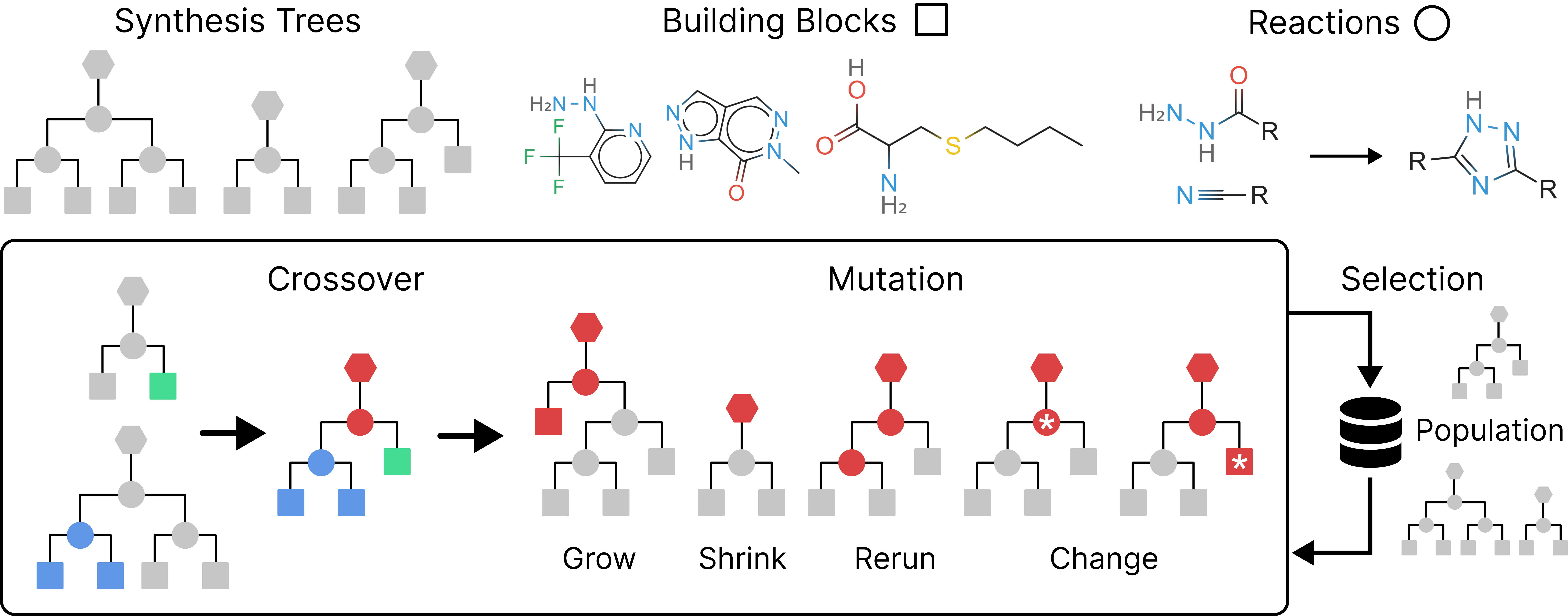}
\caption{A graphical overview of \ours{}, which operates over synthesis trees built from building blocks (squares) and reaction templates (circles). Example blocks and a reaction are drawn above using SmilesDrawer \citep{smilesDrawer}.}
\label{fig:genetic-operators}
\end{figure}

Given the power of GAs and the importance of synthesis constraints, a natural step is then to consider synthesis-constrained GAs. Prior work \citep{synnet,synformer,synthesisnet} has done so by augmenting an unconstrained GA with an ML model trained to project arbitrary molecules back onto synthesis space. While these methods are successful in synthesis planning and molecular design, they come with the upfront cost of training the ML model and the recurring cost of making inference calls to it. Moreover, they are reliant on the quality and generalization of the projection module, which can be difficult to train since it has to compress a combinatorially-large synthesis space. In this work, we take an alternate approach and directly embed synthesis constraints within the GA itself through custom genetic operators. More precisely, our contributions are:
\begin{enumerate}
\item A GA, \ours{}, that evolves synthesis routes directly (Figure \ref{fig:genetic-operators}) and is thereby explicitly synthesis-constrained. \ours{} is simple and ML-free, which makes it a nice baseline and subroutine for future algorithms, similar to unconstrained GAs.
\item An elegant way to enhance \ours{} through ML-guided building block  filtering, wherein a lightweight model is trained to dynamically restrict the block set depending on the optimization task. For property optimization, this leads to \oursbo{}, a Bayesian optimization algorithm that uses SynGA and block filtering in its inner loop.
\item Extensive benchmarks of \ours{} on various optimization tasks, where we show that \ours{} or its augmented versions achieve state-of-the-art performance. These include synthesizable analog search and sample-efficient property optimization, for 2D and 3D objectives. 
\end{enumerate}

%% file: sections/1_background.tex
\section{Background} \label{sec:background}

\textbf{Synthesis-aware molecular design.} Synthesizability can be incentivized through heuristics \citep{sascore}, reward design \citep{tango}, or fragmentation schemes \citep{crem,recap,brics}, but it can also be enforced by engineering the generative process itself. In this last case, a key design choice lies in how reactions are formalized. Template-based methods \citep{synnet,button2019} use a library of expert-defined reaction rules, whereas template-free methods  \citep{chemistga,moleculechef,dog} use an ML model to predict products. While both approaches have limitations and neither \textit{guarantee} synthesizability, an advantage of templates is they induce well-defined search spaces that do not rely on a black-box predictor. Recent template-based synthesis-aware algorithms include evolutionary algorithms \citep{synopsis,button2019,llmsynplanner}, tree search \citep{synthemol}, projection models \citep{synnet,chemproj,synformer,synthesisnet},  GFlowNets \citep{rxnflow, rgfn, synflownet}, flow matching \citep{3dsynthflow}, reinforcement learning \citep{synformer,pgfs,reactor}, and large language models \citep{synllama,llmsynplanner}. By constraining generation, these methods yield molecules with routes that can be plausibly executed using standard laboratory protocols. For a more comprehensive review, we refer readers to \citet{stanley2023fake}.

\textbf{Genetic algorithms.} GAs can be defined on a variety of molecular representations such as SMILES \citep{chemge}, SELFIES \citep{janus, nigam2020augmenting, selfies1,selfies2}, molecular graphs \citep{graphga, molga, rga, chemge}, and synthesis routes \citep{synnet, synformer, synthesisnet}. Although GAs sometimes surpass ML baselines \citep{molga,pmo}, the strongest results now emerge when they are paired with ML. Such hybridization can manifest in multiple roles: guiding a neural apprentice policy \citep{gegl}; boosting exploitation or exploration in GFlowNets \citep{geneticgfn}, Augmented Memory \citep{saturn}, and retrieval-augmented generation \citep{f-rag}; and optimizing acquisition functions in Bayesian optimization \citep{tripp2021fresh,gpbo}. When combined with synthesizable projection models, GAs can also navigate synthesizable space efficiently \citep{synnet,synformer,synthesisnet}.
Other works incorporate ML \emph{inside} the GA itself, through ML-guided genetic operators \citep{neuralga,rga} or adding learned selection pressures \citep{nigam2020augmenting,janus}, highlighting the rich design space for future hybrid methods.

%% file: sections/2_approach.tex
\section{Approach}

 We are interested in the space of the synthesizable molecules that can be obtained from a given set of building blocks and reactions. Formally, if $\Mcal$ is the universe of molecules, let $\Bcal \subseteq \Mcal$ be a finite subset of purchasable building blocks. In addition, let $\Rcal$ be a finite set of reaction rules, each one being a function $R \colon \{S \in 2^\Mcal \mid |S| = \arity(R)\} \to 2^\Mcal$ that
 maps a (multi)set of $\arity(R) \leq 2$ reactants to a set of possible products, assuming each reaction is unary or binary. In practice, $\Rcal$ is implemented by expert-defined SMARTS strings (templates), which we make invariant to input order by applying them to every permutation of the reactants and taking the union of the products as the final output. A reaction may return no products (i.e., $\varnothing$), if the input reactants are incompatible. It may also return multiple products, for example, by design or due to ambiguous regioselectivity.

New synthesizable molecules can be formed by iteratively applying reactions to the building blocks. A synthesis route producing a molecule $M$ can be represented as an unordered binary tree, where each node $v$ is labeled with a molecule $M_v$ and a reaction $R_v$ such that $M_{\text{root}} = M$ and:
\begin{enumerate}
    \item If $v$ is a leaf node, then $M_v \in \Bcal$.
    \item If $v$ is an internal node, then $M_v \in R_v(\{M_w \mid w \text{ is a child of } v\})$ and, implicitly, $v$ has exactly $\arity(R_v)$ children and they correspond to compatible reactants of $R_v$. Assigning an $M_v$ here is necessary for disambiguation, since $R_v$ can yield multiple products. 
\end{enumerate}
Conversely, any tree satisfying (1) and (2) can be interpreted as a valid synthesis route yielding a synthesizable molecule. Hence, there is a surjection $\Tcal \twoheadrightarrow \synspace$ from the set of synthesis trees $\Tcal$ to the space of synthesizable molecules $\synspace \subseteq \Mcal$, and we can cast search problems over $\Mcal_S$ as ones over $\Tcal$, which admits more compact and structured representations for ML. Here, we show that it is both possible and effective to directly search over $\Tcal$ using \ours{}, a simple genetic algorithm defined on synthesis trees.

\subsection{Genetic Algorithms}

Inspired by natural selection, genetic algorithms (GAs) \citep{holland1992adaptation} are a class of optimization algorithm that have been shown to be powerful at navigating chemical space. Generally, GAs work by iteratively updating a population of individuals through genetic operators that bias it towards higher fitness (by which we mean the value under the objective function). Commonly, GAs define three types of genetic operators: (1) crossover, which hybridizes two individuals to form a new one, (2) mutation, which locally perturbs the result of crossover, and (3) selection. At each step or generation, pairs of parents are sampled from the population and crossover and mutation are applied to produce offspring. Subsets of the offspring and current population are carried into the next generation as the new population. The selection operator defines the parent sampling and population update rules, often in a manner that favors fitter individuals. Algorithm \ref{alg:ga} gives the general structure of \ours{}. Further details and hyperparameters are given in Section \ref{sec:exp:setup}.

\textbf{Genetic operators.} Key to the success of any GA is the design of its genetic operators; our custom crossover and mutation operators enable \ours{} to operate directly over synthesis trees (Figure \ref{fig:genetic-operators}). Given two parent trees $T_1, T_2 \in \Tcal$, we perform crossover by enumerating their subtrees $\Scal_1, \Scal_2 \subseteq \Tcal$. Then, we sample a random $(S_1, S_2) \in \Scal_1 \times \Scal_2$ that are compatible with at least one bimolecular reaction $R \in \Rcal$ in the sense that $R(M(S_1), M(S_2)) \neq \varnothing$, where $M(T)$ is the final product of $T$. If such a pair exists, we join $(S_1, S_2)$ at a new root node by sampling a random compatible reaction and product. Our crossover is motivated by the intuition that if $S$ is a subtree of $T$, then $M(S)$ is similar to a fragment of $M(T)$. Our crossover corresponds roughly to fusing one fragment from each parent, which has been shown to be effective in synthesis-agnostic GAs \citep{graphga,molga}.

Given a tree $T \in \Tcal$, mutation randomly performs one of five operations:
\begin{itemize}
    \item \textbf{Grow.} Apply a random reaction $R \in \Rcal$ compatible with $M(T)$, and choose a random product. If $R$ is bimolecular, then we also sample a building block compatible with $M(T)$ and $R$. The root of $T$ becomes the child of the mutant tree's root. 
    \item \textbf{Shrink.} Randomly restrict to one of the subtrees rooted at the children of the root of $T$.
    \item \textbf{Rerun.} Keeping the blocks and reactions fixed, randomly reassign the intermediate products, i.e., $M_v$ for internal nodes $v$. Conceptually, we execute $T$ in a bottom-up (forward) direction but instead of selecting a single product $M_v \in \Mcal_S$ per reaction, we maintain and propagate sets of intermediates $\Mcal_v \subseteq \Mcal_S$, such that $\Mcal_v = \bigcup R_v(\{M_w \in \Mcal_w \mid  w \text{ is a child of } v\})$. This yields a set of alternate products $\Mcal_{\text{root}} - \{M(T)\}$ that can be produced using $T$. We randomly pick one of them and backtrack to resolve the intermediates leading to it. In practice, Rerun is implemented using a one-pass algorithm that streams the reassignments in random order, such that intermediates are only ever materialized on demand.
    \item \textbf{Change internal.} Randomly change the reaction assigned to an internal node to a new one that is compatible with its children. Rerun to obtain the mutant tree.   
    \item \textbf{Change leaf.} Randomly change the block assigned to a leaf  to a new one that is compatible with its parent and sibling (if any). Rerun to obtain the mutant tree.   
\end{itemize}
Grow and Shrink are picked with probability $0.125$ and the others with probability $0.25$. Grow can also be used to sample full synthesis trees by repeatedly applying it to a random building block for a random number of steps. We use this to initialize the population in \ours{} and to generate datasets for ML. Appendix \ref{app:impl-details} discusses additional implementation details that we omitted for clarity.

\textbf{Fitness functions.} A strength of GAs is their flexibility in the choice of fitness function $f$, allowing us to support both property optimization and analog search under a unified framework. For the former, the goal is to maximize a property $\rho$ of interest, so we can simply set $f = \rho$. For analog search, $f$ can be taken to be some notion of similarity to the query molecule. Although both problems can be framed as fitness maximization, a key distinction is that in property optimization, the fitness function is treated as an opaque oracle for which sample-efficiency is a priority. In contrast, in analog search, one can evaluate the fitness function trivially and the task's goal (the query molecule) is known. This difference impacts how we can interface \ours{} with ML.

\subsection{Building Block Filtering}

We propose an elegant ML complement to \ours{} that is deep block filtering. In the context of analog search, we learn a network $\pi_\theta \colon M \mapsto \Fcal_M \subseteq \Bcal$ that selects the most relevant building blocks $\Fcal_M$ to some query molecule $M$. If $|\Fcal_M| \ll |\Bcal|$, filtering can be highly effective, especially since our experiments use a catalog of almost 200k blocks. To search for an analog of $M$, \ours{} can then be run using $\Fcal_M$ instead of $\Bcal$. However, we consider an $\varepsilon$-filtered approach to account for potential errors made by $\pi_\theta$. When a block is to be sampled from a space $\Scal \subseteq \Bcal$, we instead sample from the filtered intersection $\Scal \cap \Fcal_M$ with probability $1 - \varepsilon$ (if nonempty), and the original subset $\Scal$ otherwise. We set $\varepsilon = 0.1$ in all our experiments.

Since $\Bcal$ is large and discrete, parameterizing a $\pi_\theta$ that selects from it is challenging. Prior work has approached this problem by using a diffusion model that generates fingerprints and then performing nearest neighbor searches \citep{synformer}. Instead, we frame the problem as a classification task. That is, we learn a binary classifier $\pi_\theta \colon \Mcal \times \Bcal \to (0, 1)$ that predicts whether a block can be used to produce a molecule. Then, we can filter $\Fcal_M = \{\pi_\theta(M, B) > \mu \mid B \in \Bcal\}$, for some threshold $\mu$.  Since $B$ is given as input, the model has explicit access to the structure of $\Bcal$ and does not have to learn it implicitly. This allows us to use a much smaller multilayer perceptron (MLP) model while achieving strong performance. We train $\pi_\theta$ on a dataset $\Dcal = \{(M, \Bcal_{M})\}$ of product-block(s) pairs obtained by randomly sampling millions of synthesis routes. We use the binary cross entropy loss and resample the positive set $\Bcal_M$ and negative set $\Bcal - \Bcal_M$ with equal probability, since the former is orders of magnitude smaller. Further details on architecture and training are given in Section \ref{sec:exp:bb-filtering}. Although $\pi_\theta$ is trained on exact product-block pairs, we use $\pi_\theta(M, B)$ to predict whether $B$ can produce an analog of $M$, if $M$ is unsynthesizable.

\subsection{Block Additive Models}

For property optimization, the above approach to block filtering is infeasible since generating a large dataset would be too sample-inefficient. Moreover, the tasks do not have explicit goal states, so the classification formulation is less applicable. Thus, we approach block filtering by fitting a neural additive model (NAM) \citep{nam} over a synthesis route's building blocks. Our choice is motivated by the simplicity and intrinsic interpretability of NAMs. Formally, given a property $\rho$ and product-block(s) pair $(M, \Bcal_M)$, the NAM models $\rho(M)$ using a sum of block-wise scores:
\begin{equation}\label{eq:nam-def}
    \rho_\theta(\Bcal_M) = \left(\alpha + (1 - \alpha)|\Bcal_M|^{-1}\right)\sum_{B \in \Bcal_M}s_\theta(B).
\end{equation}
Here, $s_\theta \colon \Bcal \to \R$ is an MLP and $\alpha \in [0, 1]$ is a learnable parameter that interpolates between a sum and mean. NAMs are easily interpretable in that each block $B$ is assigned a score $s_\theta(B)$ such that products formed from higher-scoring blocks will have higher predicted property scores. Assuming the NAM is reasonably accurate, we can then obtain a subset of promising blocks $\Fcal_M$ by filtering out the highest-scoring ones. 

Although some popular properties for drug discovery are roughly additive, others are complex and non-linear, and hence difficult to accurately model with NAMs \citep{levin2023computer}. To mitigate this, we first note that our NAM only needs to be accurate with respect to the \textit{relative} ranking between product scores. Thus, we train it with a pairwise ranking objective \citep{ranknet} instead of a regression loss. Second, we couple the NAM with a more powerful predictor that filters the samples post-hoc from \ours{}. The predictor can correct errors made by the NAM, and conversely, the NAM imposes a prior on the building block space that allows for more targeted exploration, playing an analogous role to that of a generative model. We find this is sufficient for obtaining state-of-the-art results in our experiments, though future work could explore NAMs with higher-order terms or attribution methods to improve the filter's expressivity.

\subsection{\oursbo{}}

These components are then integrated in a broader Bayesian optimization algorithm, which we call \oursbo{} (Algorithm \ref{alg:gbo}). At each step, we use SynGA with NAM filtering to maximize an acquisition function under a Gaussian process (GP) surrogate. The most fit candidates from this inner loop are evaluated under the true oracle, and the outer loop continues until the oracle budget is consumed. The NAM and GP are also periodically refitted as new samples are discovered. \oursbo{} runs \ours{} as a subroutine for roughly $100\times$ more iterations than the standard version of \ours{}, but this does not incur prohibitive cost due to the lightweight nature of \ours{} and its parallelizability. Further details are given in Appendix \ref{app:syngpbo}.

%% file: sections/3_experiments.tex
\section{Experiments}\label{sec:exp:experiments}

\subsection{Setup}\label{sec:exp:setup}

\textbf{Building blocks.} We start with 211,220 molecules from the Enamine Building Blocks catalog (US Stock, Oct.~2023) \citep{enamine} processed by \citet{chemproj}. We further discard deuterated compounds and those containing elements other than B, Br, C, Cl, F, H, I, N, O, P, S, Se, Si (e.g., organometallics). Then, the remaining molecules are sanitized and standardized using RDKit \citep{rdkit}. Lastly, removing duplicates and blocks unsupported by any reaction template leaves our final set of 196,907 building blocks.

\textbf{Reactions.} We use the reaction set from \citet{synnet}, which comprises 91 uni- or bi-molecular reaction templates compiled from \citet{hartenfeller2011} and \citet{button2019}. 

\textbf{Genetic algorithm.} We use a population size of 500, offspring size of 5, crossover rate $r_{\text{cross}} = 0.8$, mutation rate $r_{\text{mut}} = 0.5$, and elitist selection (Algorithm \ref{alg:ga}). Parents are sampled with probability proportional to their inverse rank, which is a simple approximation to the quantile-based sampling scheme used by MolGA (Appendix \ref{app:inv-rank-sampling}). To focus on small molecules, we cap all synthesis routes to at most 5 steps (internal nodes) and all products to a generous upper-bound weight of 1000 Da. 

\textbf{Fingerprints.} For ML modeling and similarity calculations, we use \textit{count} Morgan fingerprints of radius 2 by default, due to their greater specificity compared to \textit{binary} fingerprints, which can fail to discriminate between substructure repetitions. The Tanimoto similarity between fingerprints $\vx$ and $\vy$ in both cases is  $||\!\min(\vx, \vy)||_1/||\!\max(\vx, \vy)||_1$, where $\min$ and $\max$ are applied elementwise. We use 4096-dim.~fingerprints for the analog search fitness function and other similarity computations.

\subsection{Synthesizable Analog Search}

\begin{table}
\centering
\caption{Ablation of different building block filters for \ours{} on the validation set and 100 test molecules sampled from ChEMBL. We compare no filtering (\textbf{None}), a similarity heuristic (\textbf{Sim}), an MLP, and an MLP trained with hard negative mining to enhance precision (\textbf{MLP + Mine}).}
\begin{tabular}{lccccccc}
\toprule 
Filter & AUPRC & AUROC & RR & Morgan & Scaffold & Gobbi & Subset \\
\midrule 
None & $-$ & $-$     & 0.00 & 0.459 & 0.526 & 0.400 & 196,907 \\
Sim  & 0.217 & 0.970 & 0.06 & 0.625 & 0.634 & 0.515 & 9892 \\
MLP  & 0.212 &\textbf{0.999} & \textbf{0.22} & \textbf{0.721} & \textbf{0.724} & \textbf{0.635} & 117  \\
MLP + Mine  & \textbf{0.764} & \textbf{0.999} & 0.20 & 0.664 & 0.671 & 0.570 & 184 \\
\bottomrule
\end{tabular}
\label{tab:chembl_ablations}
\end{table}

\begin{table}
\centering
\caption{Average similarity scores between 1k molecules from ChEMBL and their proposed analogs. Results for SynNet and ChemProjector are taken from \citet{chemproj}. SynthesisNet and SynFormer results were reproduced with their default parameters, and we use the non-MCMC version ($\tau$ in their paper) for SynthesisNet due to compute limitations. }
\begin{tabular}{lcccccc}
\toprule 
 Method & Valid & RR & Morgan & Scaffold & Gobbi & Time  \\
\midrule 
SynNet  & 0.850 & 0.054  & 0.427 & 0.417 & 0.268 & $-$ \\
SynthesisNet  & \textbf{1.000} & 0.070 & 0.543 & 0.530 & 0.452 & $-$ \\
ChemProjector & 0.988 & 0.133 & 0.598 & 0.587  & 0.557 & $-$ \\
SynFormer  & 0.998  & 0.190 & 0.668 & 0.667 & \textbf{0.635} & \textbf{80} m \\
\midrule
\ours{} (Sim) & \textbf{1.000} & 0.064 & 0.631 & 0.638 & 0.534 & $-$ \\
\ours{} (MLP) & \textbf{1.000} & \textbf{0.196} & \textbf{0.711} & \textbf{0.694} & 0.623 & 250 m \\
\bottomrule
\end{tabular}
\label{tab:chembl_analog}
\end{table}

\subsubsection{Block Filtering} \label{sec:exp:bb-filtering}

We begin by exploring various building block filtering models for analog search. To do so, we consider the smaller-scale task of generating analogs for 100 random molecules drawn from ChEMBL \citep{chembl}. This was originally proposed in \citet{synnet} as a challenging task for assessing their model's ability to generalize to ``unreachable'' queries. Since sample-efficiency is not a primary concern in analog search, we run our GAs with a large initial population of 5k and a total budget of 10k oracle calls. To directly optimize for the evaluation metrics from \citet{chemproj}, we set the fitness function to $0.9 \cdot \text{Morgan} + 0.1 \cdot \text{Murcko}$ for the ChEMBL tasks, which are defined shortly later. For each query $M$, the most fit individual is taken as the proposed analog $A$ for further evaluation, leading to 100 query-analog pairs.  

The first two rows of Table \ref{tab:chembl_ablations} are ML-free approaches. \textbf{None} is the base \ours{}, and \textbf{Sim}\ selects all building blocks with count fingerprints $\vb$ such that $||\!\min(\vb, \vq)||_1 / ||\vb||_1 > 0.5$, where $\vq$ is the query fingerprint. Intuitively, we threshold on the fraction of local structures in the building block that are present in the query. For metrics, the reconstruction rate \textbf{RR} is the fraction of cases where $M = A$. \textbf{Morgan}, \textbf{Scaffold}, and \textbf{Gobbi} are the average similarity between $M$ and $A$ under different metrics, namely, the Tanimoto similarity between the Morgan \textit{bit} fingerprints of the pair and their Murcko scaffolds, and the dice similarity of their pharmacophore fingerprints \citep{gobbi1998}. \textbf{Subset} is the average size of the restricted block subset $|\Fcal_M|$. Surprisingly, the simple similarity  heuristic improves performance significantly and reduces the building blocks by orders of magnitude.

Inspired by this, we train a small fingerprint-based MLP model for filtering (Appendix \ref{app:filter-training}). We generate a dataset by randomly sampling synthesis routes until 10M unique products are found, and hold out 10k of them for validation. Across the validation set, the per-example \textbf{AUROC} and \textbf{AUPRC}, with respect to the $\Bcal$-wise filter scores and the binary labels, are averaged and reported in Table \ref{tab:chembl_ablations}. The MLP filter obtains strong performance on the validation set and retrieves better analogs on the test set, using a score cutoff of $0.5$. We further characterize the contributions of the GA and filter in Appendix \ref{app:rand-over-filter}, and in Appendix \ref{app:mining}, we describe how the model's precision can be increased substantially using hard negative mining  \citep{robinson2021contrastive} (\textbf{MLP + Mine}) but with degraded performance on ChEMBL. 

\subsubsection{Comparisons Against Baselines}

We benchmark \ours{} on the 1k molecule ChEMBL task from \citet{chemproj} in Table~\ref{tab:chembl_analog}. For baselines, we consider SynNet \citep{synnet}, SynthesisNet \citep{synthesisnet}, ChemProjector \citep{chemproj}, and SynFormer \citep{synformer}, which are ML models that decode a query directly into synthesis routes, represented as actions in a Markov decision process or a postfix string. The search space of \ours{} is most comparable to that of ChemProjector. SynNet and SynthesisNet use the same 91 reaction templates but an older and smaller catalog of 147k Enamine building blocks. In contrast, SynFormer uses an expanded 115 template set which includes trimolecular reactions.

Despite its smaller search space, \ours{} (MLP) achieves competitive performance in both reconstruction and analog search. In particular, some methods produce a small fraction of invalid routes (\textbf{Valid}), whereas \ours{} will always yield valid routes by design of its genetic operators. However, \ours{} is over $3\times$ slower than SynFormer (see \textbf{Time}, although our analysis has limitations discussed in Appendix \ref{app:runtime}). This is expected since amortized methods benefit from little to no searching during inference in exchange for a relatively larger model and expensive training stage. In contrast, \ours{} is more lightweight but relies on a more expensive search during inference. Hence, while \ours{} is less efficient, it is easier to adopt out of the box on new building blocks and reaction sets. Furthermore, we are able to directly optimize for arbitrary notions of chemical similarity. We discuss the strengths and weaknesses of both methods further in Appendix \ref{app:dds10_zinc}, including some extended metrics on in-distribution query sets where we find SynFormer outperforms \ours{}.

\begin{table}[t]
\centering
\caption{Projection of $N$ query molecules designed by generative models on 6 tasks. If $y$ and $y'$ are the scores of the query and analog molecules respectively, then $\Delta = y' - y$. We report the mean and standard deviation across queries, and the methods' runtimes in the header. The Valid column pertains to SynFormer and we omit it for \ours{} since it always achieves perfect validity.}
\begin{adjustbox}{max width=\textwidth}
\begin{tabular}{lccc@{\hskip 7 pt}cc@{\hskip 7 pt}c}
\toprule 
& & & \multicolumn{2}{c}{SynFormer (70 m)} & \multicolumn{2}{c}{\ours{} (MLP) (180 m)} \\[0.5ex]
 Task & $N$ & Valid & Sim. & $\Delta$ ($\uparrow$)  & Sim.  & $\Delta$ ($\uparrow$) \\
\midrule
ALDH1     & 230 
& 1.000 & $0.457          \pm 0.173$ & $\phantom{\minus} 0.118          \pm 1.190$ 
        & $\textbf{0.536} \pm 0.119$ & $\phantom{\minus} \textbf{0.302} \pm 1.134$ \\
ESR\_ant & 203 
& 1.000 & $0.553          \pm 0.151$ & $\phantom{\minus} 0.002          \pm 0.832$ 
        & $\textbf{0.644} \pm 0.134$ & $\phantom{\minus} \textbf{0.231} \pm 0.803$  \\
TP53      & 232 
& 1.000 & $0.590          \pm 0.173$ & $\phantom{\minus} 0.290          \pm 0.660$   
        & $\textbf{0.630} \pm 0.144$ & $\phantom{\minus} \textbf{0.359} \pm 0.574$  \\
\midrule 
O.~MPO & 46 
& 1.000 & $0.406          \pm 0.163$ & $\minus \textbf{0.157} \pm 0.157$ 
        & $\textbf{0.468} \pm 0.155$ & $\minus 0.162          \pm 0.152$ \\
P.~MPO & 41 
& 0.976  & $0.503          \pm 0.158$ & $\minus 0.173          \pm 0.195$ 
         & $\textbf{0.566} \pm 0.116$ & $\minus \textbf{0.156} \pm 0.197$  \\
S.~Hop    & 35 
& 1.000 & $0.523          \pm 0.112$ & $\minus 0.360          \pm 0.166$ 
        & $\textbf{0.594} \pm 0.068$ & $\minus \textbf{0.351} \pm 0.136$  \\
\bottomrule
\end{tabular}
\end{adjustbox}
\label{tab:project_sbdd_gdd}
\end{table}

\subsubsection{Projecting Structure-Based and Goal-Directed Molecular Designs}\label{sec:exp:project-designs}

Many state-of-the-art generative models are synthesis-agnostic and often propose unsynthesizable molecules \citep{gao2020synthesizability}. Commonly, SAscore \citep{sascore} is used to justify the synthetic accessibility of a model's samples. However, it has limitations as a heuristic and, ideally, we not only want to know if a molecule can be made but also how to make it. A nice use case of synthesis models such as \ours{} is then to ``project'' arbitrary designs back onto synthesis route space. Even if the original molecule is already synthesizable, doing so can uncover alternatives that are cheaper or more experimentally feasible to synthesize. In the following, we perform synthesizable hit expansions around molecules produced by structure-based and goal-directed generative models and quantify how their docking and property scores change as a result.

For structure-based design, we sample ligands binding to three receptors (ALDH1, ESR\_ant, TP53) from LIT-PCBA \citep{litpcba}, and for goal-directed design, molecules that optimize for three property oracles (Osimertinib MPO, Perindopril MPO, scaffold hop) from GuacaMol \citep{guacamol}. For ligand design, we use the \textit{negative} QuickVina2 \citep{quickvina2} docking score, so that all properties are to be maximized. Following \citet{chemproj}, for each receptor, ligands are produced using Pocket2Mol \citep{pocket2mol}, and for each GuacaMol oracle, we filter the molecules produced by \citet{gao2020synthesizability} (using three methods \citep{graphga,chemge,smileslstm} adapted by GuacaMol) that are unsynthesizable under ASKCOS \citep{askcos}. We noticed clusters of highly similar molecules in our query set, particularly in the GuacaMol tasks where high-scoring molecules are by definition those similar to some reference molecule and, hence, to each other. Consequently, we greedily cluster our dataset using a similarity cutoff of $0.7$, and retain the highest-scoring representative from each cluster.

For each query, we run \ours{} (MLP) as in Section \ref{sec:exp:bb-filtering} but optimizing only Morgan \textit{count} similarity. We dock or evaluate the five most fit analogs and retain the one with the best property score. In Table \ref{tab:project_sbdd_gdd}, we report the average similarity between the queries and analogs as well as their average difference in property score. We compare against SynFormer, using their default sampling parameters and also dropping queries that fail to decode into valid analogs for metric computation. Overall, \ours{} finds better analogs in terms of both similarity and score preservation or improvement. 

\subsection{De novo Synthesis-Aware Property Optimization}

\subsubsection{Block Additive Models} \label{sec:exp:nam}

We first ablate the utility of NAMs in the data-limited regime in Appendix \ref{app:nam}. On small datasets, we find NAMs are able to achieve good test correlation and bias products towards higher scores through building block filtering. However, by coupling NAMs with a more accurate GP predictor, we are able to surpass the performance of both components in isolation, motivating \oursbo{}'s design. 

\begin{table}[t!]
\centering
\caption{Sum of the top-10 AUC scores over the PMO suite. Results are taken from their respective papers, except MolGA, REINVENT, and SynNet are taken from \citet{geneticgfn}. We average over 5 seeds, except $f$-RAG and SynthesisNet use only 3 and 1 seeds, respectively. Separate tables for SynthesisNet and SynFlowNet are given since they were assessed only on 13 and 2 PMO tasks, instead of the full 22. Task-wise results are given in Appendix \ref{app:pmo}.}
\begin{minipage}{0.49\textwidth}
\centering
\begin{tabular}{lcc}
\toprule 
Method & Synthesis & AUC \\ 
\midrule 
$f$-RAG & \xmark & 16.301 \\ 
GPBO & \xmark & 16.304 \\ 
Genetic GFN & \xmark & 16.078 \\ 
MolGA & \xmark & 15.686 \\ 
REINVENT & \xmark & 15.003 \\ 
SynNet & \cmark & 12.610 \\ 
\midrule 
\ours{} & \cmark & 13.366 \\ 
\oursbo{} & \cmark & 16.426 \\
\bottomrule
\end{tabular}
\end{minipage}
\begin{minipage}{0.5\textwidth}
\begin{tabular}{lcc}
\toprule 
Method & Synthesis & AUC \\ 
\midrule 
SynthesisNet & \cmark & 7.906 \\ 
\ours{} & \cmark & 7.836 \\ 
\oursbo{} & \cmark & 9.332 \\ 
\bottomrule
\toprule 
Method & Synthesis & AUC \\ 
\midrule 
SynFlowNet & \cmark & 1.576 \\ 
\ours{} & \cmark & 1.842 \\ 
\oursbo{} & \cmark & 1.905 \\ 
\bottomrule
\end{tabular}
\end{minipage}
\label{tab:pmo}
\end{table}

\subsubsection{Comparisons Against Baselines: PMO} \label{sec:exp:pmo}

We run \ours{} on the Practical Molecular Optimization (PMO) benchmark \citep{pmo}, implemented by the Therapeutics Data Commons (TDC) \citep{tdc}. On the PMO benchmark, algorithms optimize against a suite of 23 tasks within 10k oracle calls. Their suggested metric is the area under the curve (AUC) of the top-10 molecules, normalized to $[0, 1]$. However, we remove the Valsartan SMARTS task for reasons discussed in Appendix \ref{app:pmo}, among other details. For baselines, we use $f$-RAG \citep{f-rag}, Genetic GFN \citep{geneticgfn}, and GPBO \citep{tripp2021fresh,gpbo}, synthesis-agnostic state-of-the-art algorithms that use MolGA as a component. $f$-RAG and Genetic GFN couple GraphGA with retrieval-augmented generation and GFlowNets, 
while GPBO uses MolGA to optimize an acquisition function in Bayesian optimization. We also report REINVENT \citep{reinvent} and MolGA \citep{molga}, which were top-performing at the time of publication of PMO. For a synthesis-aware baseline, we use SynNet \citep{synnet} which couples a synthesis projection model with a fingerprint-based GA. Table \ref{tab:pmo} provides our results along with more recent synthesis models, SynthesisNet \citep{synthesisnet} and SynFlowNet \citep{synflownet}, that were only assessed on subsets of PMO. \ours{} is competitive with the synthesis-aware algorithms on PMO, but lags behind the synthesis-agnostic algorithms. This may reflect the more constrained nature of \ours{}'s search space, which is a specific subset of (predicted) synthesizable molecules. To bridge the gap, we turn \ours{} into a model-based algorithm \oursbo{}, inspired by the significant improvement of GPBO over MolGA (Appendix \ref{app:syngpbo}). \oursbo{} attains state-of-the-art performance on PMO, being competitive with or outperforming even the top unconstrained algorithms. 

\begin{table}[t]
\centering
\caption{Vina docking scores of the top-100 diverse modes on the LIT-PCBA dataset, averaged over all receptors (Mean).  We show the scores for ALDH1, ESR\_ant, and TP53 as examples, with the full receptor-wise results  in Appendix \ref{app:docking}. Results for baselines are taken from \citet{rxnflow} and \citet{3dsynthflow}. For 3DSynthFlow, we note that its prior preprint version attained a score of $\minus 12.25 \pm 0.43$ on ALDH1. We report the mean and standard deviation over 4 seeds.}
\begin{tabular}{lcr@{\hspace{2mm}}r@{\hspace{2mm}}r@{\hspace{2mm}}r}
\toprule
Method & Calls & \multicolumn{1}{c}{ALDH1} & \multicolumn{1}{c}{ESR\_ant} & \multicolumn{1}{c}{TP53} & \multicolumn{1}{c}{{}$\,$Mean} \\
\midrule
SynNet & \multirow{6}{*}{64000} & $\minus8.81 \pm 0.21$ & $\minus8.52 \pm 0.16$ & $\minus5.34 \pm 0.23$ & $\minus8.22$ \\
BBAR &  & $\minus10.06 \pm 0.14$ & $\minus9.92 \pm 0.05$ & $\minus7.05 \pm 0.09$ & $\minus9.36$ \\
SynFlowNet &  & $\minus10.69 \pm 0.09$ & $\minus10.27 \pm 0.04$ & $\minus7.90 \pm 0.10$ & $\minus9.99$ \\
RGFN &  & $\minus9.93 \pm 0.11$ & $\minus9.72 \pm 0.14$ & $\minus7.07 \pm 0.06$ & $\minus9.20$ \\
RxnFlow &  & $\minus11.26 \pm 0.07$ & $\minus10.77 \pm 0.04$ & $\minus8.09 \pm 0.06$ & $\minus10.45$ \\
3DSynthFlow &  & $\minus11.82 \pm 0.04$ & $\minus11.23 \pm 0.08$ & $\minus8.41 \pm 0.17$ & $\minus10.88$ \\
\midrule
\ours{} & \multirow{2}{*}{16000} & $\minus11.97 \pm 0.04$ & $\minus11.07 \pm 0.20$ & $\minus8.23 \pm 0.14$ & $\minus10.80$ \\
\oursbo{} &  & $\minus\textbf{12.36} \pm 0.04$ & $\minus\textbf{11.68} \pm 0.21$ & $\minus\textbf{8.51} \pm 0.20$ & $\minus\textbf{11.11}$ \\
\bottomrule
\end{tabular}
\label{tab:docking}
\end{table}

\subsubsection{Comparisons Against Baselines: Docking}

To test \ours{} on 3D objectives, we  optimize for the UniDock \citep{unidock} Vina docking scores of LIT-PCBA receptors \citep{litpcba}, following \citet{rxnflow}. To prevent reward hacking, we use the average of the QED \citep{qed} and \textit{normalized} Vina score ($-0.1 \cdot \text{Vina}$) as our fitness function, and set a cap of 50 heavy atoms. 
We collect the top diverse modes, by filtering samples by QED $>0.5$ and greedily clustering with a similarity threshold of 0.5, and report their average docking scores in Table \ref{tab:docking}. As baselines, we consider a variety of synthesis-aware methods: SynNet \citep{synnet}, BBAR \citep{bbar}, GFlowNets  (SynFlowNet, RxnFlow, RGFN) \citep{synflownet,rxnflow,rgfn}, and 3DSynthFlow \citep{3dsynthflow}. Surprisingly, \ours{} attains better docking scores than all baselines except 3DSynthFlow with only a quarter of the oracle calls. This highlights the sample-efficiency and effectiveness of GAs. We note that 3DSynthFlow jointly designs the synthesis route and binding pose and similarly incorporating 3D information into \ours{} (e.g., as in \citet{rga}) could be promising for future work. However, we proceed by augmenting \ours{} into the model-based version \oursbo{}, in line with our approach for PMO, to attain the best docking scores overall. Further details are given in Appendix \ref{app:docking}.

%% file: sections/4_discussion.tex
\section{Conclusion} \label{sec:discussion}

We propose \ours{}, a simple synthesis-constrained GA that operates directly on synthesis routes. Within a unified framework of fitness maximization, we demonstrate the effectiveness of our method at synthesizable analog search and property optimization. \ours{} is further enhanced by ML through a lightweight  building block filter, which manifests as a classifier trained on millions of synthesis routes for analog search and an interpretable block-additive model for sample-efficient property optimization. The latter leads to the model-based variant \oursbo{}, which achieves state-of-the-art performance on both the PMO benchmark and docking tasks. However, we note that this is just one possibility, and we expect that \ours{} can be readily hybridized with ML in many other ways. We provide an extended outlook in Appendix \ref{app:outlook}.

\section*{Reproducibility Statement}

Hardware specifications are provided in Appendix \ref{app:reproducing} along with our source code, which provides documentation for preparing data and running experiments.

%% file: sections/appendix.tex
\section{Genetic Algorithm} \label{app:ga}

\subsection{Pseudocode}

In the following, the product of a synthesis tree $T \in \Tcal$ is denoted $M(T) \in \Mcal_S$. \ours{} uses elitist selection where only the fittest among the parents and offspring are retained, and is run until a budget of fitness evaluations is fully consumed.

\begin{algorithm}[H]
\caption{Pseudocode for \ours{}.}
\label{alg:ga}
\begin{algorithmic}[1]
\Require{Initial size $n_0$, population size $n$, offspring size $m$, crossover rate $r_{\text{cross}}$, mutation rate $r_{\text{mut}}$, budget $B$, a fitness function $f \colon \Mcal \to \R$.}
\State Sample $\Pcal \subseteq \Tcal$ of size $n_0$
\State $\Hcal \gets \{M(T) \to f(M(T)) \mid T \in \Pcal \}$ \Comment{keep track of unique fitness evaluations}
\While{$|\Hcal| < B$}
    \State $\Ocal \gets \varnothing$
    \For{$m$ repeats}
        \State Sample $T_1, T_2 \in \Pcal$ without replacement
        \If{$\texttt{rand}() < r_{\text{cross}}$}    
            \State $T \gets \texttt{crossover}(T_1, T_2)$
            \If{$\texttt{rand}() < r_{\text{mut}}$}  
                \State $T \gets \texttt{mutate}(T)$
            \EndIf
        \Else
            \State $T \gets \texttt{mutate}(T_1)$
        \EndIf
        \If{$T \neq \texttt{None}$ and $M(T) \notin \Hcal$ and $|\Hcal| < B$}
            \State $\Hcal[M(T)] \gets f(M(T))$
            \State $\Ocal \gets \Ocal \cup \{T\}$
        \EndIf
    \EndFor
    \State $\Pcal \gets$ the $n$ fittest individuals from $\Pcal \cup \Ocal$ 
\EndWhile
\State \Return $\Hcal$
\end{algorithmic}
\end{algorithm}

\subsection{Implementation Details} \label{app:impl-details}

\textbf{SMARTS.} We leverage the fact that the reaction templates are implemented as SMARTS strings to improve the efficiency of \ours{}. At a high level, a SMARTS string matches pattern(s) in the input molecules and defines a transformation over them to yield the product(s). The reaction will proceed with one or more products if and only if the input reactants contain the specified patterns. Concretely, a bimolecular SMARTS reaction is of the syntax \texttt{S1.S2>}\texttt{>P}, where \texttt{S1} and \texttt{S2} encode molecular substructures. The reaction will proceed if and only if one of the input molecules contains \texttt{S1} and the other \texttt{S2}. Thus, we precompute the necessary substructure matches on the base building blocks and cache them for products during runtime. In doing so, we can efficiently infer whether two molecules can react and which blocks are compatible with a given reaction (or vice versa). This strategy, however, is unlikely to scale to large libraries, although most template-based approaches use relatively small ($\sim$100) template sets to our knowledge.    

\textbf{Edge cases.} There are a number of cases in which crossover and mutation may fail. For example, a Grow operation may produce a molecule over 1000 Da, or crossover may fail to find two internal nodes that can be linked by a reaction. Depending on the edge case, we employ one of two strategies to handle it: (1) imposing boundary conditions to prevent invalidating operations from being taken, and (2) retrying the same (random) operation up to 10 times, as in GraphGA \citep{graphga}. 

\textbf{Parallelization.} Fortunately, GAs are highly amenable to parallelization. In particular, sampling the initial population, crossover and mutation for parent pairs, and evaluating the fitness function over offspring can all be implemented in a parallel manner. Benchmarks that require running multiple GA trials can also be parallelized. We leverage this in our implementation.   

\subsection{Inverse-rank sampling} \label{app:inv-rank-sampling}

MolGA \citep{molga} samples its mating pool through independent repetitions of the following: first sample $u \sim \mathcal{U}[-3, 0]$ and then sample uniformly from the top $\varepsilon = 10^u$ fraction of the population. To accommodate different population sizes $n$, suppose we instead sample $u \sim \mathcal{U}[-\log_{10}(n), 0]$. Then a change of variables shows that $\varepsilon$ has the density 
\[
    p(\varepsilon) = \frac{1}{\log_{10}(n)} \cdot \left|\frac{d}{d\varepsilon}\log_{10}(\varepsilon)\right| = \frac{1}{Z \varepsilon},
\]
supported on $[\frac{1}{n}, 1]$, for some normalization constant $Z$. Now for $1 \leq k \leq n$, the probability $p_k$ of sampling the $k$-th most fit individual is
\[
    p_k = \int_{1/n}^1 p(k \,|\, \varepsilon) p(\varepsilon)\, d\varepsilon, \quad \text{where } \;  p(k\,|\,\varepsilon) = \begin{cases}
        1/\lfloor{n\varepsilon}\rfloor, & \text{if } \varepsilon \geq k/n, \\
        0, & \text{otherwise.}
    \end{cases}
\]
Since $n$ is large and $1/\lfloor{n\varepsilon}\rfloor \approx 1 / n\varepsilon$, we can approximate that 
\[
p_k \approx \int_{k/n}^1 \frac{1}{Zn\varepsilon^2}\,d\varepsilon \propto \frac{1}{k} - \frac{1}{n} \approx \frac{1}{k}.
\]
Hence, the sampling strategy used in MolGA can be well-approximated by sampling proportionally to each individual's inverse rank. We use inverse-rank sampling for \ours{} due to its simplicity, especially when sampling without replacement. Lastly, we note that the more general distribution 
\[
p_k \propto \frac{1}{k + \lambda n},
\]
has been proposed in prior work \citep{geneticgfn}, although its mathematical connection to MolGA was not made explicit. Here, setting $\lambda = 0$ recovers inverse-rank sampling.

\subsection{Mutation Ablations}

Mutation probabilities are set proportionally to an assigned weight for each of the five actions. In our experiments, we assign a weight of 1 to Grow and Shrink, but 2 to Rerun, Change Internal (CI), and Change Leaf (CL) operation probabilities (i.e., making them twice as likely), since we expected the latter to produce more local perturbations. As a sensitivity ablation, we explore various action weights on two tasks from the PMO benchmark, following the setup in Section \ref{sec:exp:pmo}. Table \ref{tab:mut-ablate} shows that \ours{} is robust across multiple settings. We acknowledge that our ablation may not necessarily extend to other property functions, though hyperparameter tuning on the full PMO suite would also likely be overfitting.

\begin{table}[t]
\caption{Sum of the top-10 AUC scores for JNK3 and Osimertinib MPO. The last row corresponds to our current hyperparameters. We report the mean and standard deviation over 5 seeds.}
\label{tab:mut-ablate}
\centering
\begin{tabular}{cccccc}
\toprule 
Grow & Shrink & Rerun & CI & CL & AUC \\ 
\midrule 
0 & 1 & 1 & 1 & 1 & $1.465 \pm 0.068$ \\ 
1 & 0 & 1 & 1 & 1 & $1.465 \pm 0.076$ \\ 
1 & 1 & 0 & 1 & 1 & $1.496 \pm 0.052$ \\ 
1 & 1 & 1 & 0 & 1 & $1.470 \pm 0.067$ \\ 
1 & 1 & 1 & 1 & 0 & $1.451 \pm 0.104$ \\ 
1 & 1 & 1 & 1 & 1 & $1.457 \pm 0.044$ \\ 
\midrule
1 & 1 & 2 & 2 & 2 & $1.504 \pm 0.131$ \\ 
\bottomrule 
\end{tabular}
\end{table}

\section{Synthesizable Analog Search} \label{app:analog-search}

\subsection{Classifier Modelling} \label{app:filter-training}

We use a five-layer MLP of width 256 that takes as input $[\vq, \vb, \min(\vq, \vb)] \in \N_0^{3d}$, where $d = 2048$ and $\vq$ and $\vb$ are the Morgan count fingerprints of the query and building block, respectively. We use GELU activations and batch normalization. The MLP has 1.8M parameters but 1.6M of them are allocated to the first layer. The MLP is trained for 500k steps using the Adam \citep{adam} optimizer with learning rate $5 \times 10^{-4}$, and batch size 1024. We opt for an MLP for its simplicity and efficiency; the success of the fingerprint-similarity heuristic suggested that such a network could work well in the first place. We leave exploring more sophisticated architectures for future work.  

\subsection{Random Search Over Filtered Building Blocks} \label{app:rand-over-filter}

We quantify the contribution of \ours{} on the 100 molecule ChEMBL task in Table \ref{tab:chembl_filtonly_ablation}. Instead of running \ours, we sample an additional 5k synthesis routes (\textbf{Random}) under both the base and MLP-filtered building blocks. Our results suggest that, while filtering alone works reasonably well for analog search, it couples synergistically with \ours{} to produce even stronger performance.   

\begin{table}[t]
\centering
\caption{Ablation of \ours{} versus pure random search on base and MLP-filtered building blocks.}
\begin{tabular}{llccccc}
\toprule 
Method & Filter & RR & Morgan & Scaffold & Gobbi  \\
\midrule 
\multirow{2}{*}{Random}
& None & 0.00 & 0.277 & 0.364 & 0.288 \\
& MLP  & 0.14 & 0.619 & 0.629 & 0.534  \\
\midrule
\multirow{2}{*}{\ours{}}
& None & 0.00 & 0.459 & 0.526 & 0.400    \\
& MLP  & 0.22 & 0.721 & 0.724 & 0.635   \\
\bottomrule
\end{tabular}
\label{tab:chembl_filtonly_ablation}
\end{table}

\subsection{Hard-Negative Mining} \label{app:mining}

To improve the MLP filter's precision, we explore hard-negative mining from the contrastive learning literature \citep{robinson2021contrastive}. Given a molecule $M$ and building blocks $\Bcal_M \subseteq \Bcal$ that can produce it, we draw negative samples uniformly from $\Bcal - \Bcal_M$. Since $|\Bcal_M| \ll |\Bcal|$, its complement includes many blocks that are highly dissimilar to those in $\Bcal_M$, i.e., ``easy" negatives. Thus, we obtain more targeted negative examples by precomputing the 100 most similar blocks $\Ncal(B)$ to each block $B$. Then, after selecting a positive example $B_1 \in \Bcal_M$,  we sample the negative example from $\Ncal(B_1) - \Bcal_M$ with probability 0.5, and $\Bcal - \Bcal_M$ otherwise. Negative mining (\textbf{MLP + Mine} in Table \ref{tab:chembl_ablations})  
significantly improves the model's precision on the validation set, but decreases performance on ChEMBL. We attribute this to two potential reasons: (1) test molecules in ChEMBL may be out-of-distribution since the model is only ever shown ``reachable'' examples, and (2) our negative set may contain false negatives, as we can never know with certainty that a given block \textit{cannot} produce a given molecule (only that certain blocks do). This underscores the limitations behind using performance on our classification task as a direct indicator of filter quality. 

\subsection{Runtime Benchmarking} \label{app:runtime}

Experiments were run on an NVIDIA RTX A6000 GPU and a 64-core AMD Ryzen Threadripper PRO 3995WX processor. For SynFormer, we use 12 workers since too many resulted in the GPU going out of memory. For \ours{}, we use 100 workers parallelized across the batch dimension, so that each query is run with 1 worker. In general, we expect SynFormer to benefit more from better GPU compute since it requires multiple inference calls with a large ML model, in contrast to \ours{} which is predominantly CPU-bound. 

\textbf{Limitations.} Our runtime metrics in Table \ref{tab:chembl_analog} should only be taken as a rough estimate. For future work, a more careful analysis could explore different environments and inputs over multiple trials. Also of note is that \ours{} and SynFormer are research projects, whose codebases are not written with maximal efficiency in mind. It is likely that the efficiency of both methods can be improved through better engineering.

\begin{table}[t]
\centering
\caption{Extended comparison of \ours{} against SynFormer on ChEMBL and their proposed analogs. We run \ours{} for 3.3k iterations to roughly tie the runtime.}
\begin{tabular}{lcccccc}
\toprule 
 Method & Valid & RR & Morgan & Scaffold & Gobbi & Time  \\
\midrule 
SynFormer  & 0.998  & 0.190 & 0.668 & 0.667 & \textbf{0.635} & 80m \\
\ours{} (MLP, 10k) & \textbf{1.000} & \textbf{0.196} & \textbf{0.711} & \textbf{0.694} & 0.623 & 250m \\
\ours{} (MLP, 3.3k) & \textbf{1.000} & 0.175 & 0.690 & 0.675 & 0.609 & 70m \\
\bottomrule
\end{tabular}
\label{tab:chembl_tied_budget}
\end{table}
 
\begin{table}[t]
\centering
\caption{Extended comparisons of \ours{} against baselines on three datasets. ChemProjector results on DDS-10 and ZINC were obtained using the same parameters as SynFormer. We also report the average diversity within the top 5 analogs for each query.}
\begin{tabular}{llcccccc}
\toprule 
 Dataset & Method & Valid & RR & Morgan & Scaffold & Gobbi & Diversity \\
\midrule 
\multirow{3}{*}{DDS-10} 
& ChemProjector & 0.988 & 0.657 & 0.882 & 0.881 & 0.853 & 0.272 \\
& SynFormer & \textbf{1.000} & \textbf{0.858} & \textbf{0.963} & \textbf{0.966} & \textbf{0.949} & 0.243 \\ 
& \ours{} (MLP) & \textbf{1.000} & 0.707 & 0.930 & 0.924 & 0.869 & 0.272 \\ 
\midrule 
\multirow{3}{*}{ZINC}
& ChemProjector & 0.992 & 0.379 & 0.744 & 0.757 & 0.686 & 0.323 \\
& SynFormer & 0.999 & \textbf{0.558} & \textbf{0.841} & \textbf{0.845} & \textbf{0.806} & 0.290 \\ 
& \ours{} (MLP) & \textbf{1.000} & 0.459 & 0.828 & 0.826 & 0.723 & 0.280 \\
\midrule 
\multirow{3}{*}{ChEMBL}  
& ChemProjector & 0.988 & 0.133 & 0.598 & 0.587  & 0.557 & $-$ \\
& SynFormer & 0.998 & 0.190 & 0.668 & 0.667 & \textbf{0.635} & 0.343 \\
& \ours{} (MLP) & \textbf{1.000} & \textbf{0.196} & \textbf{0.711} & \textbf{0.694} & 0.623 & 0.246 \\ \bottomrule
\end{tabular}
\label{tab:dds10_zinc}
\end{table}

\subsection{Extended Comparisons Against Baselines} \label{app:dds10_zinc}

\textbf{Budget Tying.} In Table \ref{tab:chembl_tied_budget}, we compare \ours{} against SynFormer but attempt to tie the wall clock time. To do so, we run SynGA for 3.3k oracle calls to roughly match the original 3$\times$ slowdown of SynGA to SynFormer. The initial population size is also scaled down proportionally and all other hyperparameters are unchanged. We find that even with a third of the budget, SynGA exhibits strong (though slightly degraded) performance on analog search.

\textbf{Additional Test Sets.} In Table \ref{tab:dds10_zinc}, we compare \ours{} against the top two baselines from Table \ref{tab:chembl_analog} on two additional datasets: 1k molecules randomly sampled from the Enamine Discovery Diversity Set (DDS-10) \citep{dds10} and ZINC \citep{zinc}. \ours{} consistently outperforms ChemProjector, but SynFormer outperforms \ours{} on DDS-10 and ZINC. For the similarity metrics (Morgan and Scaffold) that we optimize for, the improvement is relatively modest ($\sim$0.03), but the reconstruction rate of SynFormer is considerably higher ($>$10\%). SynFormer has been extensively trained for molecular reconstruction (i.e., retrosynthesis), whereas \ours{} has only been trained on a block relevance task. Finding the precise synthesis route that produces a molecule is a needle-in-a-haystack task that may be challenging for GAs that directly search over synthesis space (we filter the top 1000 blocks, but this is still a very large search space). In contrast, ``projection'' is a smoother optimization landscape that can be more effectively navigated by GAs.  

The high reconstruction and similarity scores on DDS-10 and ZINC also suggests that both datasets are highly similar to the search spaces of all methods, in contrast to datasets like ChEMBL and the ML-designed molecules from Section \ref{sec:exp:project-designs}. DDS-10 is particularly ``reachable'', with both \ours{} and SynFormer achieving $>$0.9 average similarity, which aligns with DDS-10 and both methods' building blocks coming from Enamine (SynFormer also incorporates Enamine reaction templates). We hypothesize that out-of-distribution datasets pose problems to amortized methods similar to our findings with negative mining in Appendix \ref{app:mining}, which \ours{} can overcome through its extensive search process. Given the prior discussion, a promising direction is then to couple \ours{} with ML projection or retrosynthesis models to make up for the weaknesses of both methods. We propose some concrete ideas of this in our extended outlook (Appendix \ref{app:outlook}).

\section{De novo Synthesis-Aware Property Optimization} \label{app:prop-opt}

\begin{table}[t]
\centering
\caption{Test correlation and sample score of additive models on two oracles. Coupling the NAM with a more accurate predictor (GP) improves the score of sampled molecules. The mean and standard deviation over 5 seeds is reported.}
\begin{tabular}{lc@{\hskip 7 pt}cc@{\hskip 7 pt}c}
\toprule 
&  \multicolumn{2}{c}{JNK3} & \multicolumn{2}{c}{Osimertinib MPO} \\[0.5ex] Model & Corr. & Score  & Corr.  & Score \\
\midrule
Random 
    & $-$ & $0.058 \pm 0.002$  
    & $-$ & $0.132 \pm 0.023$ \\
Oracle (Sum)  
    & $0.739 \pm 0.127$ & $0.185 \pm 0.010$
    & $0.408 \pm 0.159$ & $0.497 \pm 0.021$ \\
Oracle (Mean) 
    & $0.411 \pm 0.183$ & $0.185 \pm 0.010$ 
    & $0.444 \pm 0.106$ & $0.497 \pm 0.021$ \\  
NAM
    & $0.877 \pm 0.043$ & $0.115 \pm 0.016$ 
    & $0.647 \pm 0.070$ & $0.603 \pm 0.027$ \\
\midrule 
GP
    & $\textbf{0.959} \pm 0.017$ & $0.188 \pm 0.009$
    & $\textbf{0.847} \pm 0.055$ & $0.679 \pm 0.005$ \\
NAM + GP
    & $-$ & $\textbf{0.297} \pm 0.050$ 
    & $-$ & $\textbf{0.714} \pm 0.020$ \\
\bottomrule
\end{tabular}
\label{tab:nam_ablate}
\end{table}

\subsection{NAM Ablations} \label{app:nam}

We first explore the utility of NAMs in the data-limited regime on two property oracles, JNK3 \citep{li2018multi} and Osimertinib MPO \citep{guacamol}. For each, we run \ours{} for 1100 oracle calls and hold out the last 100 discovered molecules as our test set. We apply a random 9:1 training-validation split of the first 1000 molecules, fit a NAM on the training set (Appendix \ref{app:nam-training}), and filter the top 1000 highest-scoring building blocks under the NAM. Then, we sample 100 synthesis routes and measure their average property score (\textbf{Score}). We also measure the Spearman correlation between the predicted and true property scores over the test set; we choose this metric because our block filtering and \ours{} are rank-based methods (e.g., elitist selection depends only on rank). 

As seen in the first section of Table \ref{tab:nam_ablate}, the NAM is able to achieve good test correlation and sampling from its filtered building blocks biases products towards higher scores, compared to random sampling from an unfiltered block set (\textbf{Random}). We also report \textbf{Oracle}, which takes $s_\theta$ in Equation \ref{eq:nam-def} to be the property function and $\alpha = 1$ (Sum) or $\alpha = 0$ (Mean). This can be thought of an idealized model that measures how additive the property functions are. The NAM's performance can be pushed further by coupling it with a stronger predictive model. We fit a Gaussian process (GP) to the training set (Appendix \ref{app:gp}), sample 10k synthesis routes, and use the posterior mean to select 100 products for further evaluation. Their average scores are given in the second section along with the GP's test set correlation. \textbf{NAM + GP} samples from the NAM-filtered building blocks, whereas \textbf{GP} uses just the base blocks. As expected, the GP is more accurate than the NAM, but by coupling the two we are able to surpass the performance of both components in isolation.

\subsection{NAM Modelling} \label{app:nam-training}

To implement the NAM $s_\theta$, we use a five-layer MLP of width 64 that takes as input the 2048-count Morgan fingerprint of the input block. We use GELU activations and no normalization. The NAM has 140k parameters. We train the NAM using the Adam \citep{adam} optimizer with learning rate $5 \times 10^{-4}$, batch size 50, and early stopping on the validation Spearman correlation with a 5 epoch patience. We use the RankNet \citep{ranknet} objective which computes the loss as:
\[
\Lcal(\theta) = \mathrm{BCE}\bigg(\rho_\theta(\Bcal_1) - \rho_\theta(\Bcal_2),\, \mathbb{I}\left[\rho(M_1) > \rho(M_2)\right]  \bigg),
\]
for a pair of examples $(M_1, \Bcal_1)$ and $(M_2, \Bcal_2)$, where $\rho_\theta$ is the NAM and $\rho$ is the property function and $\mathbb{I}$ is the indicator function. We average the loss over pairwise combinations of the batch. 

We found ranking loss to outperform the standard mean-squared-error (MSE) loss for both the NAM and NAM + GP models (Table \ref{tab:nam_ablate_loss}). The ranking objective leads to better NAM test correlation on both objectives. On Osimertinib MPO, this translates to an increase in sample scores. On JNK3, scores are marginally worse, which we hypothesize is because the JNK3 oracle is already well-approximated by additive models, as shown in Table \ref{tab:nam_ablate}. 

\begin{table}[t!]
\centering
\caption{Ablation of MSE versus ranking loss for NAM training. The mean and standard deviation over 5 seeds is reported.}
\begin{tabular}{llc@{\hskip 7 pt}cc@{\hskip 7 pt}c}
\toprule 
&  & \multicolumn{2}{c}{JNK3} & \multicolumn{2}{c}{Osimertinib MPO} \\[0.5ex] Model & Loss  & Corr. & Score  & Corr.  & Score \\
\midrule
\multirow{2}{*}{NAM} 
& MSE
& $0.864 \pm 0.035$ & $0.118 \pm 0.023$
& $0.458 \pm 0.110$ & $0.450 \pm 0.058$ \\
& Rank
& $\textbf{0.877} \pm 0.043$ & $0.115 \pm 0.016$
& $\textbf{0.647} \pm 0.070$ & $0.603 \pm 0.027$ \\
\midrule 
\multirow{2}{*}{NAM + GP} 
& MSE
& $-$ & $\textbf{0.301} \pm 0.059$
& $-$ & $0.709 \pm 0.007$ \\
& Rank
& $-$ & $0.297 \pm 0.050$
& $-$ & $\textbf{0.714} \pm 0.020$ \\
\bottomrule
\end{tabular}
\label{tab:nam_ablate_loss}
\end{table}

\subsection{Gaussian Processes} \label{app:gp}

Our Gaussian process uses 2048-count Morgan fingerprints as the features, the MinMax kernel from Gauche \citep{gauche}, and GPytorch \citep{gpytorch} for its implementation.

\subsection{\oursbo{}} \label{app:syngpbo}

Inspired by GPBO \citep{gpbo}, we convert \ours{} into a model-based variant \oursbo{}. At a high level, \oursbo{} uses \ours{} to optimize an acquisition function within a broader Bayesian optimization loop. At each step, a GP and NAM are fit to the samples discovered thus far, following Appendix \ref{app:nam-training} and \ref{app:gp}. Since GPs scale poorly with dataset size, we subset to the top 2500 samples and a random subset of 2500 other samples in practice, and to avoid repeated retrainings, we only refit the NAM every 25 steps. Then, we use \ours{} to optimize an acquisition function under the GP surrogate. Following \citet{gpbo}, we use the upper confidence bound (UCB) acquisition with $\beta \sim [0.01, 1]$ sampled logarithmically until 5000 samples are obtained, after which we set $\beta = 0$ and maximize the posterior mean. The process repeats until a budget of oracle calls is exhausted (Algorithm \ref{alg:gbo}). 

In the inner loop, we run \ours{} for 5 generations with an offspring size of 100. We use a population size of 1000 starting with 500 randomly sampled individuals and the top 1000 scoring molecules. All other parameters are kept the same as Section \ref{sec:exp:setup}. In total, \ours{} proposes $500 + 5 \cdot 100 = 1000$ \textit{new} molecules at most, of which the 10 most fit ones are evaluated by the true oracle. Hence, \ours{} proposes $100$ molecules for every molecule evaluated. In contrast, GPBO proposes roughly 1000. As noted by \citet{gpbo}, performance can likely be improved by increasing the number of outer and inner loop iterations of \oursbo{}.

\begin{algorithm}[H]
\caption{Pseudocode for \oursbo{}.}
\label{alg:gbo}
\begin{algorithmic}[1]
\Require{Proposal size $m$, budget $B$, GP $g$, NAM $s_\theta$, a fitness function $f \colon \Mcal \to \R$.}
\State $\Hcal \gets \{M(T_i) \to f(M(T_i)) \mid T_i \in \Tcal\}$ for $m$ samples  
\State $i \gets 0$
\While{$|\Hcal| < B$}
    \If{$|\Hcal| \geq 500$ and $i \equiv 0$ (mod 25)}
        \State Fit $s_\theta$ to $\Hcal$
        \State $\Fcal \gets$ top-1000 scoring blocks in $\Bcal$ under $s_\theta$
    \Else
        \State $\Fcal \gets \Bcal$
    \EndIf{}
    \State Fit $g$ to $\Hcal$, or a subset if $|\Hcal|$ is large
    \State $\Pcal_0 \gets $ top-1000 candidates in $\Hcal$ and 500 random routes
    \State $\alpha \gets \mathrm{UCB}(g, \beta)$ for $\beta \sim p(\beta)$
    \State Run SynGA with filtered blocks $\Fcal$ from initial population $\Pcal_0$ with fitness function $\alpha$ 
    \For{the $m$ fittest individuals $T$ from SynGA}
        \State $\Hcal[M(T)] \gets f(M(T))$
    \EndFor 
    \State $i \gets i + 1$ 
\EndWhile
\State \Return $\Hcal$
\end{algorithmic}
\end{algorithm}

\subsection{Practical Molecular Optimization Benchmark} \label{app:pmo}

The top-$k$ AUC PMO metric is formally defined as $\frac{1}{B}\sum_{t = 1}^B \bar{\rho}_{k, t}$, where $B$ is the budget and $\bar{\rho}_{k, t}$ is the average of the top $k$ oracle scores within the first $t$ samples. In \citet{pmo}, this is further estimated using the trapezoidal rule at 100 sample intervals. Tables \ref{tab:pmo_full}, \ref{tab:pmo_full2}, and \ref{tab:pmo_full3} give the expanded task-wise results of Table \ref{tab:pmo}. 
 
\textbf{PyTDC.} The original PMO implementation from \citet{pmo} used PyTDC 0.3.6 \citep{tdc}. On 0.3.7 onwards, PyTDC made a bug fix\footnote{\texttt{\url{https://github.com/mims-harvard/TDC/pull/171}}} that led to breaking changes in the Isomers, Sitagliptin MPO, and Zaleplon MPO oracles. For example, \citet{geneticgfn} reproduce PMO with PyTDC 0.4.0, and we observe a consistent increase in PMO scores in their results. Thus, care must be taken to pin the PyTDC version when using numbers from \citet{pmo}. However, we found PyTDC 0.3.6 difficult to install due to  dependency conflicts and its  pins to overly  old versions of some libraries. In the interest of using up-to-date packages, we use PyTDC 1.1.14 and avoid numbers from the original PMO paper. We also spot check 10k molecules from ZINC \citep{zinc} to confirm that there are no discrepancies between PyTDC 1.1.14 and 0.3.6 for oracles other than those mentioned above.   

\textbf{Valsartan SMARTS.} Many methods fail to optimize the Valsartan SMARTS task. This is because the oracle returns 0 if the input does not contain \texttt{CN(C=O)Cc1ccc(c2ccccc2)cc1}. That is, models are given no signal (i.e., the oracle appears constant) until they propose a molecule with one specific substructure. But a priori, without any signal or information about the task, there is no reason an algorithm should do so. For this reason, we argue Valsartan SMARTS is ill-suited for benchmarking and remove it. 

\begin{figure}[t]
\centering
\includegraphics[width=0.95\linewidth]{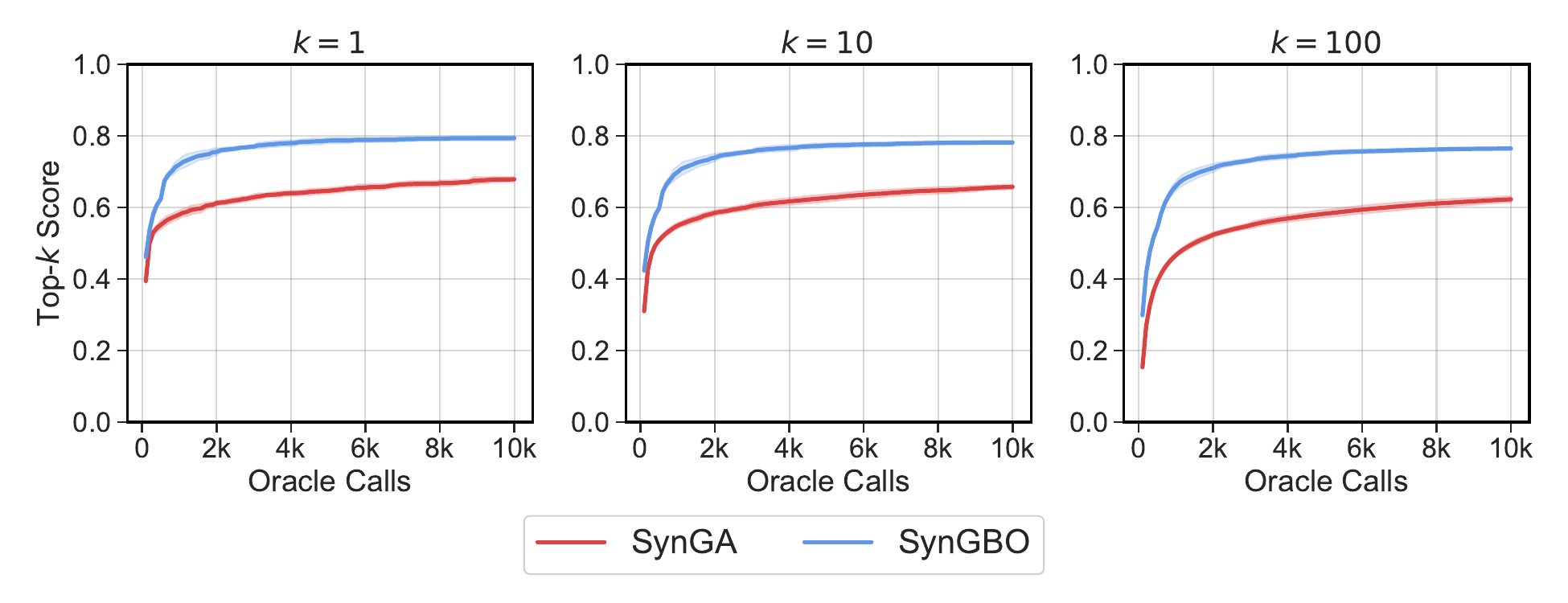}
\caption{The mean score for the top-$k$ molecules plotted over the number of oracle calls consumed, and averaged over tasks. We plot the mean over 5 seeds.}
\label{fig:pmo_curves}
\end{figure}

\textbf{Extended Results.} Table \ref{tab:pmo_more_metrics} reports other metrics for \ours{} and \oursbo{}, for completeness, and Figure \ref{fig:pmo_curves} provides some optimization curves.

\begin{table}[t]
\centering
\caption{Number of discovered modes (i.e., docking score < $\minus$10, QED > 0.5, similarity threshold of 0.5) for the ALDH1 task. Results for baselines are taken from \citet{3dsynthflow}. We report the mean and standard deviation over 4 seeds.}
\begin{tabular}{lccc}
\toprule 
Method	& 1k Calls	& 5k Calls &	10k Calls \\
\midrule 
RxnFlow & $4.5 \pm 2.1$ & $26.5 \pm 7.8$ & $73.5 \pm 33.2$ \\
3DSynthFlow & $18.5 \pm 14.8$ & $112.0 \pm 94.8$ & $326.5 \pm 316.1$ \\
\midrule 
\ours{} & $32.5 \pm 1.5$ & $144.2 \pm 16.2$ & $241.5 \pm 29.0$ \\
\oursbo{} & $50.5 \pm 18.5$ & $171.8 \pm 36.8$ & $182.0 \pm 30.9$ \\
\bottomrule
\end{tabular}
\label{tab:docking_modes}
\end{table}

\subsection{LIT-PCBA Docking Benchmark} \label{app:docking}

To better leverage the GPU batching of UniDock, we increase the offspring size of \ours{} to 100 and the proposal size of \oursbo{} to 20. Since the benchmark emphasizes diversity, we further increase the population size of \ours{} to 5000 and initial population size to 1000. Table \ref{tab:docking_vina_full} gives the expanded receptor-wise results of Table \ref{tab:docking}.

\begin{figure}[p]
    \centering
    \includegraphics[width=0.95\textwidth]{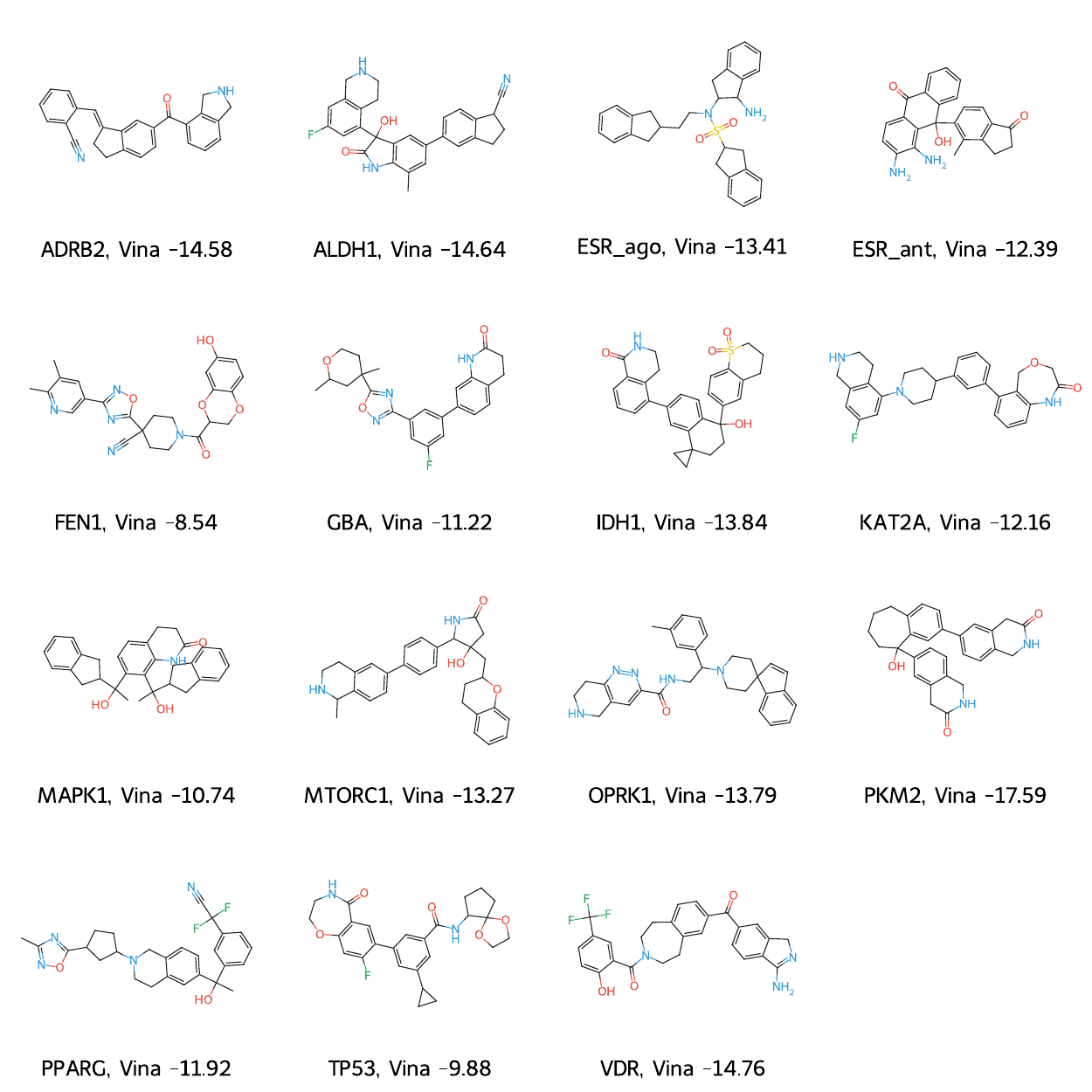}
    \bigskip
    \caption{An example ligand proposed by \ours{} for each receptor.}
    \label{fig:synga-dock-top1}
\end{figure}

\begin{figure}[p]
    \centering
    \includegraphics[width=0.95\textwidth]{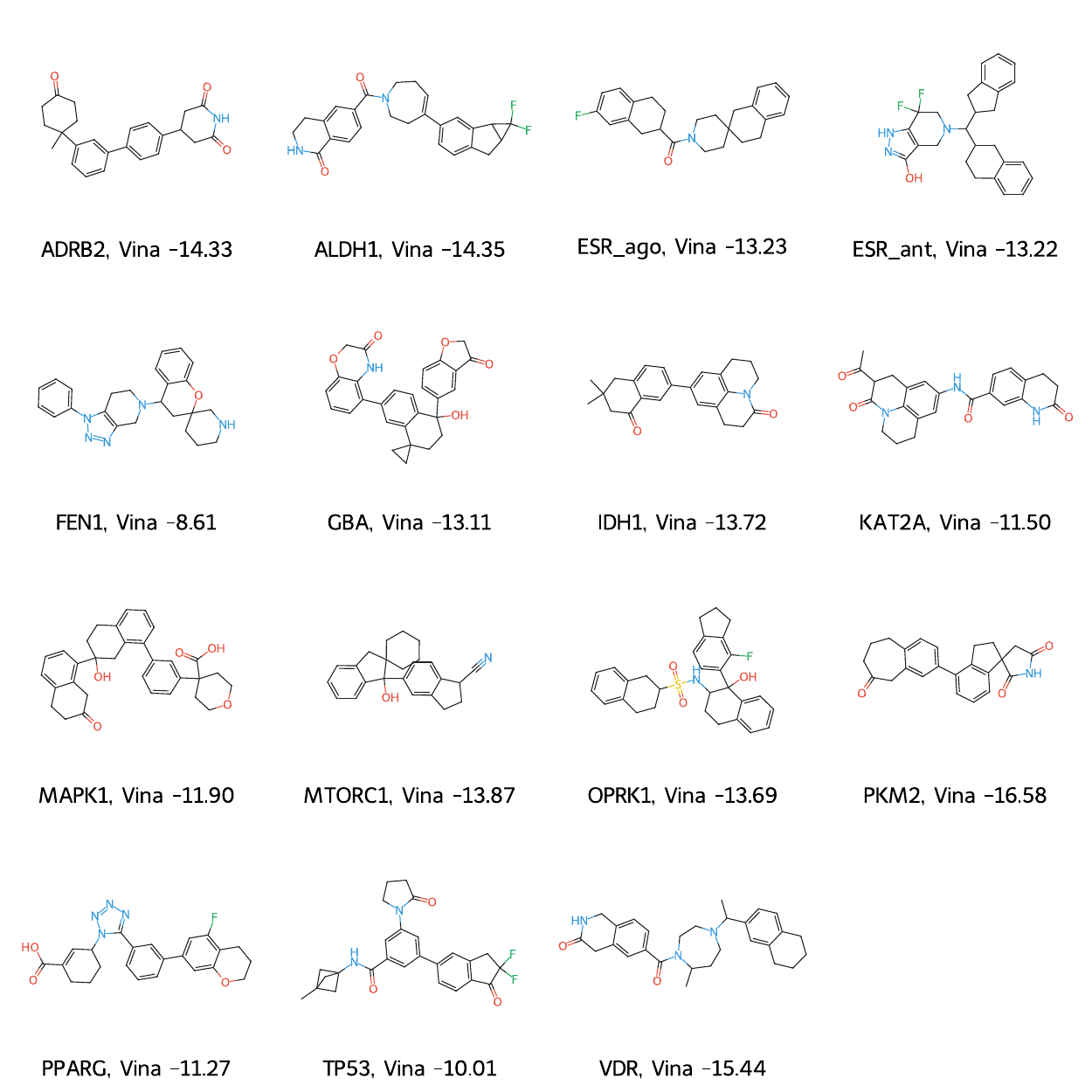}
    \bigskip
    \caption{An example ligand proposed by \oursbo{} for each receptor.}
    \label{fig:syngpbo-dock-top1}
\end{figure}

\textbf{Extended Results.} Table \ref{tab:docking_eff_full} reports the average ligand efficiency (Vina score normalized by heavy atom count) of the top modes. For a diversity-focused metric, we report the number of discovered ALDH1 modes as a function of oracle calls in Table \ref{tab:docking_modes}. \ours{} and \oursbo{} perform more diverse exploration than baselines in the early stages of optimization but fall behind 3DSynthFlow afterwards. We attribute this to the exploitative nature of GAs whose goal is property optimization, whereas GFlowNets and CGFlow instead aim to sample proportionally to the property or reward. Example ligands proposed by \ours{} and \oursbo{} are displayed in Figures \ref{fig:synga-dock-top1} and \ref{fig:syngpbo-dock-top1}, which we obtain from the top mode across all seeds that passes the Tartarus \citep{tartarus} filters and whose ring systems all appear in ChEMBL at least 5 times.\footnote{\url{https://github.com/PatWalters/useful_rdkit_utils}} For future work, many of these filters can be  applied on the block- or GA-level to minimize the number of post-hoc rejected samples.

Finally, we perform a preliminary ablation aiming to match the search space of 3DSynthFlow. We consider both the 1.2M and 300k block sets mentioned in their work, which are obtained from Enamine's comprehensive and in-stock catalogs, respectively. For the former, only the current-day catalog version is available and, since it has grown to 1.6M blocks, we randomly downsample it to 1.2M blocks. In both cases, we use the same 38 bimolecular reaction templates, 2 maximum reaction steps, and 40 maximum atoms. We report our results on two receptors in Table~\ref{tab:docking_ablation}. All hyperparameters are kept the same, except we filter the top 10k blocks when using the 1.2M catalog set to accommodate for the larger block space. We find the general trend holds: using only a quarter of the oracle calls, \oursbo{} outperforms or competes with 3DSynthFlow, and \ours{} outperforms or competes with all other baselines.

\begin{table}[t]
\centering
\caption{Vina docking scores of the top-100 diverse modes on two LIT-PCBA receptors, over various search spaces. Only the top 3 baselines are shown for conciseness. We report the mean and standard deviation over 4 seeds.}
\label{tab:docking_ablation}
\begin{tabular}{llcc}
\toprule
Method & Search Space & ESR\_ant & TP53 \\
\midrule
SynFlowNet & & $\minus 10.27 \pm 0.04$ & $\minus 7.90 \pm 0.10$ \\
RxnFlow & & $\minus 10.77 \pm 0.04$ & $\minus 8.09 \pm 0.06$ \\
3DSynthFlow & & $\minus 11.23 \pm 0.08$ & $\minus 8.41 \pm 0.17$ \\
\midrule
\ours{} & Ours & $\minus 11.07 \pm 0.20$ & $\minus 8.23 \pm 0.14$ \\
\ours{} & 3DSynthFlow (1.2M) & $\minus 10.92 \pm 0.15$ & $\minus 8.17 \pm 0.25$ \\
\ours{} & 3DSynthFlow (300k) & $\minus 10.89 \pm 0.13$ & $\minus 8.24 \pm 0.13$ \\
\midrule
\oursbo{} & Ours & $\minus 11.68 \pm 0.21$ & $\minus 8.51 \pm 0.20$ \\
\oursbo{} & 3DSynthFlow (1.2M) & $\minus 11.39 \pm 0.12$ & $\minus 8.72 \pm 0.33$ \\
\oursbo{} & 3DSynthFlow (300k) & $\minus 11.53 \pm 0.04$ & $\minus 8.45 \pm 0.41$ \\
\bottomrule
\end{tabular}
\end{table}

\section{Extended Outlook}\label{app:outlook}

\textbf{Limitations.} \ours{} inherits the general limitations of template-based approaches to synthesis. For example, our templates do not \textit{guarantee} synthesizability, nor do they consider reaction conditions, stereochemistry, yield, or cost. Any fixed template library also necessarily restricts exploration to a biased subset of synthesizable space. This can be mitigated by enlarging the template set, though a naïve extension may degrade efficiency and robustness. In addition, the tasks considered in this work are single-objective (or scalarized) whereas real-world molecular design is highly multi-objective. Fortunately, multi-objective GAs such as NSGA-II \citep{nsgaii} can be easily integrated with our proposed genetic operators. Many of the tasks are also synthesis-agnostic and we treat synthesizability as an additional dimension that contextualizes our results.
Evaluating \ours{} on benchmarks such as Tartarus \citep{tartarus} that attempt to penalize ``unreasonable" samples could enrich our comparisons with synthesis-agnostic baselines for future work. 

\textbf{Future directions.} While we present \ours{} as a standalone work,  our hope is that \ours{} can also serve as a building block for future ML algorithms. For analog search, one prospective direction could be to refine the outputs of synthesis models like SynFormer \citep{synformer} by running \ours{} briefly, and to potentially even improve the model by finetuning on the better analogs. For property optimization, \ours{} can be used to boost exploitation or exploration in generative models. Augmenting \ours{} with ML may also be promising, such as using a 3D network to enhance the genetic operators for docking tasks. Finally, we note that \ours{} is just one synthesis-constrained GA, and future work can look into exploring the rich design space of genetic operators.

\section{Reproducibility} \label{app:reproducing}

\textbf{Compute.} All experiments were run on a single NVIDIA RTX A6000 GPU and a 64-core AMD Ryzen Threadripper PRO 3995WX processor. The compute used for analog search is discussed in Appendix \ref{app:runtime} and runtimes are given in Table \ref{tab:chembl_analog}. Training the MLP block filter took $\sim$4 hours. For PMO, \ours{} took $\sim$20 min per trial with 5 workers and \oursbo{} took $\sim$6 hours per trial with 20 workers. However, we ran many trials concurrently on the same machine, so these times are likely inflated. For the docking experiments, \ours{} took ${\sim} 3$ hours with 50 workers and \oursbo{} took ${\sim} 4$ hours with 20 workers. We ran trials sequentially and we found that computing the docking scores was a significant portion of the runtime.  

\textbf{Code.} \url{https://github.com/alstonlo/synga}.

\newpage

\input{tables/pmo}

\begin{table}[t]
\centering
\caption{Extended metrics on the PMO benchmark: the mean score and AUC score for the top-$k$ molecules summed across tasks, as well as their average diversity. We report the mean and standard deviation over 5 seeds. The minor discrepancy in top-10 AUC scores with Table \ref{tab:pmo} is due to rounding. Here, we average the metrics per seed and round as a final step, whereas Table \ref{tab:pmo} computes the metrics task-wise, rounds, and then sums.}
\begin{tabular}{llrr}
\toprule 
 & Metric & \multicolumn{1}{c}{\ours{}} & \multicolumn{1}{c}{\oursbo{}} \\
\midrule 
Top-1
& Mean & $14.927 \pm 0.164$ & $17.460 \pm 0.142$ \\
\midrule 
\multirow{3}{*}{Top-10}  
& Mean & $14.464 \pm 0.152$ & $17.195 \pm 0.115$ \\ 
& AUC  & $13.369 \pm 0.175$ & $16.425 \pm 0.116$ \\ 
& Diversity & $ 0.514 \pm 0.008$ & $ 0.388 \pm 0.011$ \\
\midrule 
\multirow{3}{*}{Top-100}
& Mean & $13.699 \pm 0.199$ & $16.824 \pm 0.095$ \\ 
& AUC  & $12.211 \pm 0.187$ & $15.856 \pm 0.113$ \\ 
& Diversity & $ 0.613 \pm 0.014$ & $ 0.462 \pm 0.010$ \\
\bottomrule
\end{tabular}
\label{tab:pmo_more_metrics}
\end{table}

\clearpage   

\input{tables/litpcba}

%% file: tables/pmo.tex
\begin{table}[p]
\centering
\caption{Task-wise results of Table \ref{tab:pmo}.}
\begin{tabular}{lccccc}
\toprule
Oracle & $f$-RAG  & GPBO  & G.~GFN  & \ours{} & \oursbo{} \\
Synthesis & \xmark & \xmark & \xmark & \cmark & \cmark \\ 
\midrule
Albu.~Sim. & \pmocell{0.977 $\pm$ 0.002} & \pmocell{0.964 $\pm$ 0.050} & \pmocell{0.949 $\pm$ 0.010} & \pmocell{0.649 $\pm$ 0.058} & \pmocell{0.947 $\pm$ 0.024} \\
Amlo.~MPO & \pmocell{0.749 $\pm$ 0.019} & \pmocell{0.720 $\pm$ 0.061} & \pmocell{0.761 $\pm$ 0.019} & \pmocell{0.573 $\pm$ 0.019} & \pmocell{0.670 $\pm$ 0.088} \\
Cele.~Redisc. & \pmocell{0.778 $\pm$ 0.007} & \pmocell{0.860 $\pm$ 0.002} & \pmocell{0.802 $\pm$ 0.029} & \pmocell{0.494 $\pm$ 0.063} & \pmocell{0.856 $\pm$ 0.013} \\
Deco Hop & \pmocell{0.936 $\pm$ 0.011} & \pmocell{0.672 $\pm$ 0.118} & \pmocell{0.733 $\pm$ 0.109} & \pmocell{0.629 $\pm$ 0.014} & \pmocell{0.831 $\pm$ 0.039} \\
DRD2 & \pmocell{0.992 $\pm$ 0.000} & \pmocell{0.902 $\pm$ 0.117} & \pmocell{0.974 $\pm$ 0.006} & \pmocell{0.976 $\pm$ 0.006} & \pmocell{0.981 $\pm$ 0.010} \\
Fexo.~MPO & \pmocell{0.856 $\pm$ 0.016} & \pmocell{0.806 $\pm$ 0.006} & \pmocell{0.856 $\pm$ 0.039} & \pmocell{0.773 $\pm$ 0.018} & \pmocell{0.833 $\pm$ 0.018} \\
GSK3$\beta$ & \pmocell{0.969 $\pm$ 0.003} & \pmocell{0.877 $\pm$ 0.055} & \pmocell{0.881 $\pm$ 0.042} & \pmocell{0.866 $\pm$ 0.072} & \pmocell{0.924 $\pm$ 0.027} \\
Isom.~C7H8. & \pmocell{0.955 $\pm$ 0.008} & \pmocell{0.911 $\pm$ 0.031} & \pmocell{0.969 $\pm$ 0.003} & \pmocell{0.840 $\pm$ 0.016} & \pmocell{0.975 $\pm$ 0.006} \\
Isom.~C9H10. & \pmocell{0.850 $\pm$ 0.005} & \pmocell{0.828 $\pm$ 0.126} & \pmocell{0.897 $\pm$ 0.007} & \pmocell{0.707 $\pm$ 0.040} & \pmocell{0.875 $\pm$ 0.013} \\
JNK3 & \pmocell{0.904 $\pm$ 0.004} & \pmocell{0.785 $\pm$ 0.072} & \pmocell{0.764 $\pm$ 0.069} & \pmocell{0.683 $\pm$ 0.132} & \pmocell{0.910 $\pm$ 0.021} \\
Median 1 & \pmocell{0.340 $\pm$ 0.007} & \pmocell{0.415 $\pm$ 0.001} & \pmocell{0.379 $\pm$ 0.010} & \pmocell{0.254 $\pm$ 0.017} & \pmocell{0.357 $\pm$ 0.001} \\
Median 2 & \pmocell{0.323 $\pm$ 0.005} & \pmocell{0.408 $\pm$ 0.003} & \pmocell{0.294 $\pm$ 0.007} & \pmocell{0.226 $\pm$ 0.009} & \pmocell{0.349 $\pm$ 0.001} \\
Mest.~Sim. & \pmocell{0.671 $\pm$ 0.021} & \pmocell{0.930 $\pm$ 0.106} & \pmocell{0.708 $\pm$ 0.057} & \pmocell{0.480 $\pm$ 0.008} & \pmocell{0.759 $\pm$ 0.023} \\
Osim.~MPO & \pmocell{0.866 $\pm$ 0.009} & \pmocell{0.833 $\pm$ 0.011} & \pmocell{0.860 $\pm$ 0.008} & \pmocell{0.820 $\pm$ 0.003} & \pmocell{0.856 $\pm$ 0.024} \\
Peri.~MPO & \pmocell{0.681 $\pm$ 0.017} & \pmocell{0.651 $\pm$ 0.030} & \pmocell{0.595 $\pm$ 0.014} & \pmocell{0.556 $\pm$ 0.032} & \pmocell{0.774 $\pm$ 0.006} \\
QED & \pmocell{0.939 $\pm$ 0.001} & \pmocell{0.947 $\pm$ 0.000} & \pmocell{0.942 $\pm$ 0.000} & \pmocell{0.938 $\pm$ 0.001} & \pmocell{0.940 $\pm$ 0.002} \\
Rano.~MPO & \pmocell{0.820 $\pm$ 0.016} & \pmocell{0.810 $\pm$ 0.011} & \pmocell{0.819 $\pm$ 0.018} & \pmocell{0.802 $\pm$ 0.009} & \pmocell{0.839 $\pm$ 0.016} \\
Scaffold Hop & \pmocell{0.576 $\pm$ 0.014} & \pmocell{0.529 $\pm$ 0.020} & \pmocell{0.615 $\pm$ 0.100} & \pmocell{0.532 $\pm$ 0.014} & \pmocell{0.541 $\pm$ 0.008} \\
Sita.~MPO & \pmocell{0.601 $\pm$ 0.011} & \pmocell{0.474 $\pm$ 0.085} & \pmocell{0.634 $\pm$ 0.039} & \pmocell{0.348 $\pm$ 0.022} & \pmocell{0.454 $\pm$ 0.074} \\
Thio.~Redisc. & \pmocell{0.584 $\pm$ 0.009} & \pmocell{0.727 $\pm$ 0.089} & \pmocell{0.583 $\pm$ 0.034} & \pmocell{0.433 $\pm$ 0.033} & \pmocell{0.647 $\pm$ 0.003} \\
Trog.~Redisc. & \pmocell{0.448 $\pm$ 0.017} & \pmocell{0.756 $\pm$ 0.141} & \pmocell{0.511 $\pm$ 0.054} & \pmocell{0.322 $\pm$ 0.013} & \pmocell{0.579 $\pm$ 0.002} \\
Zale.~MPO & \pmocell{0.486 $\pm$ 0.004} & \pmocell{0.499 $\pm$ 0.025} & \pmocell{0.552 $\pm$ 0.033} & \pmocell{0.465 $\pm$ 0.017} & \pmocell{0.529 $\pm$ 0.017} \\
\midrule
Sum & 16.301 & 16.304 & 16.078 & 13.366 & 16.426 \\
\bottomrule
\end{tabular}
\label{tab:pmo_full}
\bigskip
\caption{Task-wise results of Table \ref{tab:pmo} (continued).}
\centerline{
\begin{tabular}{lccccc}
\toprule
Oracle & REINVENT  & MolGA   & SynNet & \ours{} & \oursbo{} \\
Synthesis & \xmark & \xmark & \cmark & \cmark & \cmark \\ 
\midrule
Albu.~Sim. & \pmocell{0.881 $\pm$ 0.016} & \pmocell{0.928 $\pm$ 0.015} & \pmocell{0.568 $\pm$ 0.033} & \pmocell{0.649 $\pm$ 0.058} & \pmocell{0.947 $\pm$ 0.024} \\
Amlo.~MPO & \pmocell{0.644 $\pm$ 0.019} & \pmocell{0.740 $\pm$ 0.055} & \pmocell{0.566 $\pm$ 0.006} & \pmocell{0.573 $\pm$ 0.019} & \pmocell{0.670 $\pm$ 0.088} \\
Cele.~Redisc. & \pmocell{0.717 $\pm$ 0.027} & \pmocell{0.629 $\pm$ 0.062} & \pmocell{0.439 $\pm$ 0.035} & \pmocell{0.494 $\pm$ 0.063} & \pmocell{0.856 $\pm$ 0.013} \\
Deco Hop & \pmocell{0.662 $\pm$ 0.044} & \pmocell{0.656 $\pm$ 0.013} & \pmocell{0.635 $\pm$ 0.043} & \pmocell{0.629 $\pm$ 0.014} & \pmocell{0.831 $\pm$ 0.039} \\
DRD2 & \pmocell{0.957 $\pm$ 0.007} & \pmocell{0.950 $\pm$ 0.004} & \pmocell{0.970 $\pm$ 0.006} & \pmocell{0.976 $\pm$ 0.006} & \pmocell{0.981 $\pm$ 0.010} \\
Fexo.~MPO & \pmocell{0.781 $\pm$ 0.013} & \pmocell{0.835 $\pm$ 0.012} & \pmocell{0.750 $\pm$ 0.016} & \pmocell{0.773 $\pm$ 0.018} & \pmocell{0.833 $\pm$ 0.018} \\
GSK3$\beta$ & \pmocell{0.885 $\pm$ 0.031} & \pmocell{0.894 $\pm$ 0.025} & \pmocell{0.713 $\pm$ 0.057} & \pmocell{0.866 $\pm$ 0.072} & \pmocell{0.924 $\pm$ 0.027} \\
Isom.~C7H8. & \pmocell{0.942 $\pm$ 0.012} & \pmocell{0.926 $\pm$ 0.014} & \pmocell{0.862 $\pm$ 0.004} & \pmocell{0.840 $\pm$ 0.016} & \pmocell{0.975 $\pm$ 0.006} \\
Isom.~C9H10. & \pmocell{0.838 $\pm$ 0.030} & \pmocell{0.894 $\pm$ 0.005} & \pmocell{0.657 $\pm$ 0.030} & \pmocell{0.707 $\pm$ 0.040} & \pmocell{0.875 $\pm$ 0.013} \\
JNK3 & \pmocell{0.782 $\pm$ 0.029} & \pmocell{0.835 $\pm$ 0.040} & \pmocell{0.574 $\pm$ 0.103} & \pmocell{0.683 $\pm$ 0.132} & \pmocell{0.910 $\pm$ 0.021} \\
Median 1 & \pmocell{0.363 $\pm$ 0.011} & \pmocell{0.329 $\pm$ 0.006} & \pmocell{0.236 $\pm$ 0.015} & \pmocell{0.254 $\pm$ 0.017} & \pmocell{0.357 $\pm$ 0.001} \\
Median 2 & \pmocell{0.281 $\pm$ 0.002} & \pmocell{0.284 $\pm$ 0.035} & \pmocell{0.241 $\pm$ 0.007} & \pmocell{0.226 $\pm$ 0.009} & \pmocell{0.349 $\pm$ 0.001} \\
Mest.~Sim. & \pmocell{0.634 $\pm$ 0.042} & \pmocell{0.762 $\pm$ 0.048} & \pmocell{0.402 $\pm$ 0.017} & \pmocell{0.480 $\pm$ 0.008} & \pmocell{0.759 $\pm$ 0.023} \\
Osim.~MPO & \pmocell{0.834 $\pm$ 0.010} & \pmocell{0.853 $\pm$ 0.005} & \pmocell{0.793 $\pm$ 0.008} & \pmocell{0.820 $\pm$ 0.003} & \pmocell{0.856 $\pm$ 0.024} \\
Peri.~MPO & \pmocell{0.535 $\pm$ 0.015} & \pmocell{0.610 $\pm$ 0.038} & \pmocell{0.541 $\pm$ 0.021} & \pmocell{0.556 $\pm$ 0.032} & \pmocell{0.774 $\pm$ 0.006} \\
QED & \pmocell{0.941 $\pm$ 0.000} & \pmocell{0.941 $\pm$ 0.001} & \pmocell{0.941 $\pm$ 0.001} & \pmocell{0.938 $\pm$ 0.001} & \pmocell{0.940 $\pm$ 0.002} \\
Rano.~MPO & \pmocell{0.770 $\pm$ 0.005} & \pmocell{0.830 $\pm$ 0.010} & \pmocell{0.749 $\pm$ 0.009} & \pmocell{0.802 $\pm$ 0.009} & \pmocell{0.839 $\pm$ 0.016} \\
Scaffold Hop & \pmocell{0.551 $\pm$ 0.024} & \pmocell{0.568 $\pm$ 0.017} & \pmocell{0.506 $\pm$ 0.012} & \pmocell{0.532 $\pm$ 0.014} & \pmocell{0.541 $\pm$ 0.008} \\
Sita.~MPO & \pmocell{0.470 $\pm$ 0.041} & \pmocell{0.677 $\pm$ 0.055} & \pmocell{0.297 $\pm$ 0.033} & \pmocell{0.348 $\pm$ 0.022} & \pmocell{0.454 $\pm$ 0.074} \\
Thio.~Redisc. & \pmocell{0.544 $\pm$ 0.026} & \pmocell{0.544 $\pm$ 0.067} & \pmocell{0.397 $\pm$ 0.012} & \pmocell{0.433 $\pm$ 0.033} & \pmocell{0.647 $\pm$ 0.003} \\
Trog.~Redisc. & \pmocell{0.458 $\pm$ 0.018} & \pmocell{0.487 $\pm$ 0.024} & \pmocell{0.280 $\pm$ 0.006} & \pmocell{0.322 $\pm$ 0.013} & \pmocell{0.579 $\pm$ 0.002} \\
Zale.~MPO & \pmocell{0.533 $\pm$ 0.009} & \pmocell{0.514 $\pm$ 0.033} & \pmocell{0.493 $\pm$ 0.014} & \pmocell{0.465 $\pm$ 0.017} & \pmocell{0.529 $\pm$ 0.017} \\
\midrule 
Sum & 15.003 & 15.686 & 12.610 & 13.366 & 16.426 \\
\bottomrule
\end{tabular}
}
\label{tab:pmo_full2}
\end{table}

\clearpage

\begin{table}[t]
\centering
\caption{Task-wise results of Table \ref{tab:pmo} (continued).}
\begin{tabular}{lcccc}
\toprule
Oracle & SynthesisNet  & SynFlowNet  & \ours{} & \oursbo{} \\
Synthesis & \cmark & \cmark & \cmark & \cmark \\ 
\midrule
Amlo.~MPO & 
    \pmocell{0.608} & $-$  & \pmocell{0.573 $\pm$ 0.019} & \pmocell{0.670 $\pm$ 0.088} \\
Cele.~Redisc. &
    \pmocell{0.582} & $-$ & \pmocell{0.494 $\pm$ 0.063} & \pmocell{0.856 $\pm$ 0.013} \\
DRD2 & 
    \pmocell{0.960} & \pmocell{0.885 $\pm$ 0.027} & \pmocell{0.976 $\pm$ 0.006} & \pmocell{0.981 $\pm$ 0.010} \\
Fexo.~MPO & 
    \pmocell{0.791} & $-$ & \pmocell{0.773 $\pm$ 0.018} & \pmocell{0.833 $\pm$ 0.018}  \\
GSK3$\beta$ & 
    \pmocell{0.848} & \pmocell{0.691 $\pm$ 0.034} & \pmocell{0.866 $\pm$ 0.072} & \pmocell{0.924 $\pm$ 0.027} \\
JNK3 & 
    \pmocell{0.639} & $-$ & \pmocell{0.683 $\pm$ 0.132} & \pmocell{0.910 $\pm$ 0.021} \\
Median 1 & \pmocell{0.305} & $-$ &\pmocell{0.254 $\pm$ 0.017} & \pmocell{0.357 $\pm$ 0.001} \\
Median 2 &
    \pmocell{0.257} & $-$ & \pmocell{0.226 $\pm$ 0.009} & \pmocell{0.349 $\pm$ 0.001} \\
Osim.~MPO & 
    \pmocell{0.810} & $-$ & \pmocell{0.820 $\pm$ 0.003} & \pmocell{0.856 $\pm$ 0.024} \\
Peri.~MPO & 
    \pmocell{0.524} & $-$ & \pmocell{0.556 $\pm$ 0.032} & \pmocell{0.774 $\pm$ 0.006} \\
Rano.~MPO & 
    \pmocell{0.741} & $-$ & \pmocell{0.802 $\pm$ 0.009} & \pmocell{0.839 $\pm$ 0.016} \\
Sita.~MPO & 
    \pmocell{0.313} & $-$ & \pmocell{0.348 $\pm$ 0.022} & \pmocell{0.454 $\pm$ 0.074} \\
Zale.~MPO & 
    \pmocell{0.528} & $-$ & \pmocell{0.465 $\pm$ 0.017} & \pmocell{0.529 $\pm$ 0.017} \\
\bottomrule
\end{tabular}
\label{tab:pmo_full3}
\end{table}

%% file: tables/litpcba.tex
\newcommand{\mytabskip}{0.1in}

\begin{table}[p]
\centering
\caption{Vina docking scores of the top-100 diverse modes on LIT-PCBA dataset. Results for baselines are taken from \citet{rxnflow} and \citet{3dsynthflow} and use 64000 oracle calls whereas we use 16000. We report the mean and standard deviation over 4 seeds.}
{
\resizebox{\textwidth}{!}{\begin{tabular}{lr@{\hspace{2mm}}r@{\hspace{2.5mm}}r@{\hspace{2mm}}r@{\hspace{2mm}}r}
\toprule
Method & \multicolumn{1}{c}{ADRB2} & \multicolumn{1}{c}{ALDH1} & \multicolumn{1}{c}{ESR\_ago} & \multicolumn{1}{c}{ESR\_ant} & \multicolumn{1}{c}{FEN1} \\
\midrule
SynNet & $\minus8.03 \pm 0.26$ & $\minus8.81 \pm 0.21$ & $\minus8.88 \pm 0.13$ & $\minus8.52 \pm 0.16$ & $\minus6.36 \pm 0.09$ \\
BBAR & $\minus9.95 \pm 0.04$ & $\minus10.06 \pm 0.14$ & $\minus9.97 \pm 0.03$ & $\minus9.92 \pm 0.05$ & $\minus6.84 \pm 0.07$ \\
SynFlowNet & $\minus10.85 \pm 0.10$ & $\minus10.69 \pm 0.09$ & $\minus10.44 \pm 0.05$ & $\minus10.27 \pm 0.04$ & $\minus7.47 \pm 0.02$ \\
RGFN & $\minus9.84 \pm 0.21$ & $\minus9.93 \pm 0.11$ & $\minus9.99 \pm 0.11$ & $\minus9.72 \pm 0.14$ & $\minus6.92 \pm 0.06$ \\
RxnFlow & $\minus11.45 \pm 0.05$ & $\minus11.26 \pm 0.07$ & $\minus11.15 \pm 0.02$ & $\minus10.77 \pm 0.04$ & $\minus7.66 \pm 0.02$ \\
3DSynthFlow & $\minus11.97 \pm 0.12$ & $\minus11.82 \pm 0.04$ & $\minus\textbf{11.58} \pm 0.07$ & $\minus11.23 \pm 0.08$ & $\minus\textbf{7.79} \pm 0.01$ \\
\midrule
\ours{} & $\minus11.98 \pm 0.27$ & $\minus11.97 \pm 0.04$ & $\minus11.23 \pm 0.11$ & $\minus11.07 \pm 0.20$ & $\minus7.61 \pm 0.06$ \\
\oursbo{} & $\minus\textbf{12.16} \pm 0.42$ & $\minus\textbf{12.36} \pm 0.04$ & $\minus11.45 \pm 0.05$ & $\minus\textbf{11.68} \pm 0.21$ & $\minus7.71 \pm 0.12$ \\
\bottomrule
\end{tabular}}
\vspace{\mytabskip}

\resizebox{\textwidth}{!}{\begin{tabular}{lr@{\hspace{2mm}}r@{\hspace{2.5mm}}r@{\hspace{2mm}}r@{\hspace{2mm}}r}
\toprule
Method & \multicolumn{1}{c}{GBA} & \multicolumn{1}{c}{IDH1} & \multicolumn{1}{c}{KAT2A} & \multicolumn{1}{c}{MAPK1} & \multicolumn{1}{c}{MTORC1} \\
\midrule
SynNet & $\minus7.60 \pm 0.09$ & $\minus8.74 \pm 0.08$ & $\minus7.64 \pm 0.38$ & $\minus7.33 \pm 0.14$ & $\minus9.30 \pm 0.45$ \\
BBAR & $\minus8.70 \pm 0.05$ & $\minus9.84 \pm 0.09$ & $\minus8.54 \pm 0.06$ & $\minus8.49 \pm 0.07$ & $\minus10.07 \pm 0.16$ \\
SynFlowNet & $\minus9.27 \pm 0.06$ & $\minus10.40 \pm 0.08$ & $\minus9.41 \pm 0.04$ & $\minus8.92 \pm 0.05$ & $\minus10.84 \pm 0.03$ \\
RGFN & $\minus8.48 \pm 0.06$ & $\minus9.49 \pm 0.13$ & $\minus8.53 \pm 0.11$ & $\minus8.22 \pm 0.15$ & $\minus9.89 \pm 0.06$ \\
RxnFlow & $\minus9.62 \pm 0.04$ & $\minus10.95 \pm 0.05$ & $\minus9.73 \pm 0.03$ & $\minus9.30 \pm 0.01$ & $\minus11.39 \pm 0.09$ \\
3DSynthFlow & $\minus9.90 \pm 0.14$ & $\minus11.28 \pm 0.15$ & $\minus\textbf{10.17} \pm 0.37$ & $\minus9.61 \pm 0.11$ & $\minus11.91 \pm 0.01$ \\
\midrule
\ours{} & $\minus9.65 \pm 0.16$ & $\minus11.27 \pm 0.21$ & $\minus10.08 \pm 0.11$ & $\minus9.38 \pm 0.20$ & $\minus12.04 \pm 0.45$ \\
\oursbo{} & $\minus\textbf{10.33} \pm 0.09$ & $\minus\textbf{11.76} \pm 0.28$ & $\minus10.08 \pm 0.10$ & $\minus\textbf{9.64} \pm 0.09$ & $\minus\textbf{12.28} \pm 0.26$ \\
\bottomrule
\end{tabular}}
\vspace{\mytabskip}

\resizebox{\textwidth}{!}{\begin{tabular}{lr@{\hspace{2mm}}r@{\hspace{2.5mm}}r@{\hspace{2mm}}r@{\hspace{2mm}}r}
\toprule
Method & \multicolumn{1}{c}{OPRK1} & \multicolumn{1}{c}{PKM2} & \multicolumn{1}{c}{PPARG} & \multicolumn{1}{c}{TP53} & \multicolumn{1}{c}{VDR} \\
\midrule
SynNet & $\minus8.70 \pm 0.36$ & $\minus9.55 \pm 0.14$ & $\minus7.47 \pm 0.34$ & $\minus5.34 \pm 0.23$ & $\minus10.98 \pm 0.57$ \\
BBAR & $\minus9.84 \pm 0.10$ & $\minus11.39 \pm 0.08$ & $\minus8.69 \pm 0.10$ & $\minus7.05 \pm 0.09$ & $\minus11.07 \pm 0.04$ \\
SynFlowNet & $\minus10.34 \pm 0.07$ & $\minus11.98 \pm 0.12$ & $\minus9.40 \pm 0.05$ & $\minus7.90 \pm 0.10$ & $\minus11.62 \pm 0.13$ \\
RGFN & $\minus9.61 \pm 0.11$ & $\minus10.96 \pm 0.18$ & $\minus8.53 \pm 0.07$ & $\minus7.07 \pm 0.06$ & $\minus10.86 \pm 0.11$ \\
RxnFlow & $\minus10.84 \pm 0.03$ & $\minus12.53 \pm 0.02$ & $\minus9.73 \pm 0.02$ & $\minus8.09 \pm 0.06$ & $\minus12.30 \pm 0.07$ \\
3DSynthFlow & $\minus11.26 \pm 0.41$ & $\minus13.36 \pm 0.03$ & $\minus10.00 \pm 0.04$ & $\minus8.41 \pm 0.17$ & $\minus12.98 \pm 0.10$ \\
\midrule
\ours{} & $\minus11.25 \pm 0.19$ & $\minus13.44 \pm 0.12$ & $\minus10.00 \pm 0.15$ & $\minus8.23 \pm 0.14$ & $\minus12.77 \pm 0.21$ \\
\oursbo{} & $\minus\textbf{11.55} \pm 0.12$ & $\minus\textbf{13.65} \pm 0.09$ & $\minus\textbf{10.18} \pm 0.08$ & $\minus\textbf{8.51} \pm 0.20$ & $\minus\textbf{13.36} \pm 0.12$ \\
\bottomrule
\end{tabular}}
}
\label{tab:docking_vina_full}
\end{table}

\begin{table}[p]
\centering
\caption{Ligand efficiency of the top-100 diverse modes on the LIT-PCBA dataset. Results for baselines are taken from \citet{rxnflow} and \citet{3dsynthflow} and use 64000 oracle calls whereas we use 16000. We report the mean and standard deviation over 4 seeds.}
{
\resizebox{\textwidth}{!}{\begin{tabular}{lccccc}
\toprule
Method & \multicolumn{1}{c}{ADRB2} & \multicolumn{1}{c}{ALDH1} & \multicolumn{1}{c}{ESR\_ago} & \multicolumn{1}{c}{ESR\_ant} & \multicolumn{1}{c}{FEN1} \\
\midrule
SynNet & $0.274 \pm 0.041$ & $0.272 \pm 0.006$ & $0.317 \pm 0.005$ & $0.289 \pm 0.020$ & $0.196 \pm 0.003$ \\
BBAR & $0.412 \pm 0.006$ & $0.401 \pm 0.008$ & $0.380 \pm 0.001$ & $0.387 \pm 0.003$ & $0.257 \pm 0.003$ \\
SynFlowNet & $0.401 \pm 0.006$ & $0.380 \pm 0.007$ & $0.361 \pm 0.003$ & $0.361 \pm 0.004$ & $0.247 \pm 0.004$ \\
RGFN & $0.393 \pm 0.005$ & $0.357 \pm 0.004$ & $0.346 \pm 0.002$ & $0.344 \pm 0.002$ & $0.241 \pm 0.001$ \\
RxnFlow & $0.412 \pm 0.005$ & $0.396 \pm 0.005$ & $0.375 \pm 0.002$ & $0.380 \pm 0.004$ & $0.246 \pm 0.001$ \\
3DSynthFlow & $0.448 \pm 0.019$ & $0.395 \pm 0.006$ & $0.391 \pm 0.005$ & $0.398 \pm 0.016$ & $0.252 \pm 0.005$ \\
\midrule
\ours{} & $0.435 \pm 0.012$ & $0.411 \pm 0.010$ & $0.387 \pm 0.005$ & $0.385 \pm 0.006$ & $0.249 \pm 0.004$ \\
\oursbo{} & $\textbf{0.486} \pm 0.019$ & $\textbf{0.459} \pm 0.008$ & $\textbf{0.422} \pm 0.008$ & $\textbf{0.451} \pm 0.007$ & $\textbf{0.285} \pm 0.006$ \\
\bottomrule
\end{tabular}}
\vspace{\mytabskip}

\resizebox{\textwidth}{!}{\begin{tabular}{lccccc}
\toprule
Method & \multicolumn{1}{c}{GBA} & \multicolumn{1}{c}{IDH1} & \multicolumn{1}{c}{KAT2A} & \multicolumn{1}{c}{MAPK1} & \multicolumn{1}{c}{MTORC1} \\
\midrule
SynNet & $0.244 \pm 0.013$ & $0.281 \pm 0.016$ & $0.294 \pm 0.042$ & $0.226 \pm 0.004$ & $0.316 \pm 0.035$ \\
BBAR & $0.336 \pm 0.002$ & $0.382 \pm 0.005$ & $0.332 \pm 0.003$ & $0.320 \pm 0.002$ & $0.385 \pm 0.004$ \\
SynFlowNet & $0.330 \pm 0.004$ & $0.368 \pm 0.002$ & $0.327 \pm 0.003$ & $0.305 \pm 0.002$ & $0.368 \pm 0.002$ \\
RGFN & $0.310 \pm 0.002$ & $0.351 \pm 0.003$ & $0.310 \pm 0.003$ & $0.298 \pm 0.002$ & $0.346 \pm 0.004$ \\
RxnFlow & $0.327 \pm 0.004$ & $0.378 \pm 0.001$ & $0.330 \pm 0.001$ & $0.313 \pm 0.001$ & $0.370 \pm 0.001$ \\
3DSynthFlow & $0.340 \pm 0.018$ & $0.388 \pm 0.007$ & $0.340 \pm 0.016$ & $0.322 \pm 0.009$ & $0.387 \pm 0.008$ \\
\midrule
\ours{} & $0.327 \pm 0.002$ & $0.401 \pm 0.009$ & $0.334 \pm 0.004$ & $0.328 \pm 0.003$ & $0.377 \pm 0.006$ \\
\oursbo{} & $\textbf{0.390} \pm 0.006$ & $\textbf{0.442} \pm 0.004$ & $\textbf{0.378} \pm 0.005$ & $\textbf{0.365} \pm 0.007$ & $\textbf{0.421} \pm 0.008$ \\
\bottomrule
\end{tabular}}
\vspace{\mytabskip}

\resizebox{\textwidth}{!}{\begin{tabular}{lccccc}
\toprule
Method & \multicolumn{1}{c}{OPRK1} & \multicolumn{1}{c}{PKM2} & \multicolumn{1}{c}{PPARG} & \multicolumn{1}{c}{TP53} & \multicolumn{1}{c}{VDR} \\
\midrule
SynNet & $0.298 \pm 0.039$ & $0.296 \pm 0.005$ & $0.253 \pm 0.031$ & $0.211 \pm 0.031$ & $0.359 \pm 0.015$ \\
BBAR & $0.370 \pm 0.006$ & $0.442 \pm 0.004$ & $0.326 \pm 0.007$ & $0.288 \pm 0.005$ & $0.409 \pm 0.002$ \\
SynFlowNet & $0.359 \pm 0.004$ & $0.427 \pm 0.003$ & $0.317 \pm 0.002$ & $0.287 \pm 0.008$ & $0.393 \pm 0.003$ \\
RGFN & $0.349 \pm 0.001$ & $0.405 \pm 0.002$ & $0.307 \pm 0.002$ & $0.271 \pm 0.001$ & $0.381 \pm 0.002$ \\
RxnFlow & $0.369 \pm 0.007$ & $0.436 \pm 0.005$ & $0.319 \pm 0.002$ & $0.289 \pm 0.003$ & $0.405 \pm 0.002$ \\
3DSynthFlow & $0.389 \pm 0.006$ & $0.444 \pm 0.010$ & $0.321 \pm 0.005$ & $0.294 \pm 0.006$ & $0.416 \pm 0.009$ \\
\midrule
\ours{} & $0.383 \pm 0.006$ & $0.494 \pm 0.005$ & $0.329 \pm 0.005$ & $0.299 \pm 0.011$ & $0.415 \pm 0.010$ \\
\oursbo{} & $\textbf{0.459} \pm 0.013$ & $\textbf{0.543} \pm 0.005$ & $\textbf{0.363} \pm 0.006$ & $\textbf{0.336} \pm 0.008$ & $\textbf{0.465} \pm 0.009$ \\
\bottomrule
\end{tabular}}
}
\label{tab:docking_eff_full}
\end{table}

%% file: references.bib
@inproceedings{3dsynthflow,
title={Compositional Flows for 3D Molecule and Synthesis Pathway Co-design},
author={Tony Shen and Seonghwan Seo and Ross Irwin and Kieran Didi and Simon Olsson and Woo Youn Kim and Martin Ester},
booktitle={Forty-second International Conference on Machine Learning},
year={2025},
}

@inproceedings{adam,
  title        = {Adam: {A} Method for Stochastic Optimization},
  author       = {Diederik P. Kingma and Jimmy Ba},
  year         = {2015},
  booktitle    = {3rd International Conference on Learning Representations, {ICLR} 2015, San Diego, CA, USA, May 7-9, 2015, Conference Track Proceedings},
}

@article{aizynthfinder,
	title = {{AiZynthFinder} 4.0: developments based on learnings from 3 years of industrial application},
	volume = {16},
	issn = {1758-2946},
	doi = {10.1186/s13321-024-00860-x},
	number = {1},
	journal = {Journal of Cheminformatics},
	author = {Saigiridharan, Lakshidaa and Hassen, Alan Kai and Lai, Helen and Torren-Peraire, Paula and Engkvist, Ola and Genheden, Samuel},
	month = may,
	year = {2024},
	pages = {57},
}

@article{askcos,
  title        = {A robotic platform for flow synthesis of organic compounds informed by {AI} planning},
  author       = {Connor W. Coley  and Dale A. Thomas  and Justin A. M. Lummiss  and Jonathan N. Jaworski  and Christopher P. Breen  and Victor Schultz  and Travis Hart  and Joshua S. Fishman  and Luke Rogers  and Hanyu Gao  and Robert W. Hicklin  and Pieter P. Plehiers  and Joshua Byington  and John S. Piotti  and William H. Green  and A. John Hart  and Timothy F. Jamison  and Klavs F. Jensen},
  year         = {2019},
  journal      = {Science},
  volume       = {365},
  number       = {6453},
  pages        = {eaax1566},
  doi          = {10.1126/science.aax1566},
}

@article{bbar,
  title        = {Molecular generative model via retrosynthetically prepared chemical building block assembly},
  author       = {Seo, Seonghwan and Lim, Jaechang and Kim, Woo Youn},
  year         = {2023},
  journal      = {Adv. Sci.},
  publisher    = {Wiley},
  volume       = {10},
  number       = {8},
  pages        = {2206674},
}

@inproceedings{bengio2021flow,
  title        = {Flow Network based Generative Models for Non-Iterative Diverse Candidate Generation},
  author       = {Emmanuel Bengio and Moksh Jain and Maksym Korablyov and Doina Precup and Yoshua Bengio},
  year         = {2021},
  booktitle    = {Advances in Neural Information Processing Systems},
}

@article{bombarelli2018automatic,
  title        = {Automatic Chemical Design Using a Data-Driven Continuous Representation of Molecules},
  author       = {Gómez-Bombarelli, Rafael and Wei, Jennifer N. and Duvenaud, David and Hernández-Lobato, Jos{\'e} Miguel and Sánchez-Lengeling, Benjamín and Sheberla, Dennis and Aguilera-Iparraguirre, Jorge and Hirzel, Timothy D. and Adams, Ryan P. and Aspuru-Guzik, Alán},
  year         = {2018},
  journal      = {ACS Central Science},
  volume       = {4},
  number       = {2},
  pages        = {268--276},
  doi          = {10.1021/acscentsci.7b00572},
}

@article{brics,
  title        = {On the Art of Compiling and Using `Drug-Like' Chemical Fragment Spaces},
  author       = {Degen, Jörg and Wegscheid-Gerlach, Christof and Zaliani, Andrea and Rarey, Matthias},
  year         = {2008},
  journal      = {ChemMedChem},
  volume       = {3},
  number       = {10},
  pages        = {1503--1507},
  doi          = {10.1002/cmdc.200800178},
}

@article{button2019,
  title        = {Automated de novo molecular design by hybrid machine intelligence and rule-driven chemical synthesis},
  author       = {Button, Alexander and Merk, Daniel and Hiss, Jan A. and Schneider, Gisbert},
  year         = {2019},
  month        = {Jul},
  day          = {01},
  journal      = {Nature Machine Intelligence},
  volume       = {1},
  number       = {7},
  pages        = {307--315},
  doi          = {10.1038/s42256-019-0067-7},
}

@article{chembl,
  title        = {The {ChEMBL} Database in 2023: a drug discovery platform spanning multiple bioactivity data types and time periods},
  author       = {Zdrazil, Barbara and Felix, Eloy and Hunter, Fiona and Manners, Emma J and Blackshaw, James and Corbett, Sybilla and de Veij, Marleen and Ioannidis, Harris and Lopez, David Mendez and Mosquera, Juan F and Magarinos, Maria Paula and Bosc, Nicolas and Arcila, Ricardo and Kizilören, Tevfik and Gaulton, Anna and Bento, A Patrícia and Adasme, Melissa F and Monecke, Peter and Landrum, Gregory A and Leach, Andrew R},
  year         = {2023},
  month        = {11},
  journal      = {Nucleic Acids Research},
  volume       = {52},
  number       = {D1},
  pages        = {D1180-D1192},
  doi          = {10.1093/nar/gkad1004},
}

@article{chemcrow,
  title        = {Augmenting large language models with chemistry tools},
  author       = {M. Bran, Andres and Cox, Sam and Schilter, Oliver and Baldassari, Carlo and White, Andrew D. and Schwaller, Philippe},
  year         = {2024},
  month        = {May},
  day          = {01},
  journal      = {Nature Machine Intelligence},
  volume       = {6},
  number       = {5},
  pages        = {525--535},
  doi          = {10.1038/s42256-024-00832-8},
}

@article{chemge,
  title        = {Population-based De Novo Molecule Generation, Using Grammatical Evolution},
  author       = {Yoshikawa, Naruki and Terayama, Kei and Sumita, Masato and Homma, Teruki and Oono, Kenta and Tsuda, Koji},
  year         = {2018},
  month        = {10},
  journal      = {Chemistry Letters},
  volume       = {47},
  number       = {11},
  pages        = {1431--1434},
  doi          = {10.1246/cl.180665},
}

@article{chemistga,
  title        = {{ChemistGA}: A Chemical Synthesizable Accessible Molecular Generation Algorithm for Real-World Drug Discovery},
  author       = {Wang, Jike and Wang, Xiaorui and Sun, Huiyong and Wang, Mingyang and Zeng, Yundian and Jiang, Dejun and Wu, Zhenxing and Liu, Zeyi and Liao, Ben and Yao, Xiaojun and Hsieh, Chang-Yu and Cao, Dongsheng and Chen, Xi and Hou, Tingjun},
  year         = {2022},
  journal      = {Journal of Medicinal Chemistry},
  volume       = {65},
  number       = {18},
  pages        = {12482--12496},
  doi          = {10.1021/acs.jmedchem.2c01179},
}

@inproceedings{chemproj,
  title        = {Projecting Molecules into Synthesizable Chemical Spaces},
  author       = {Shitong Luo and Wenhao Gao and Zuofan Wu and Jian Peng and Connor W. Coley and Jianzhu Ma},
  year         = {2024},
  booktitle    = {Forty-first International Conference on Machine Learning},
}

@article{crem,
  title        = {{CReM}: chemically reasonable mutations framework for structure generation},
  author       = {Polishchuk, Pavel},
  year         = {2020},
  month        = {Apr},
  day          = {22},
  journal      = {Journal of Cheminformatics},
  volume       = {12},
  number       = {1},
  pages        = {28},
  doi          = {10.1186/s13321-020-00431-w},
}

@inproceedings{dog,
  title        = {Barking up the right tree: an approach to search over molecule synthesis {DAGs}},
  author       = {Bradshaw, John and Paige, Brooks and Kusner, Matt J and Segler, Marwin and Hern\'{a}ndez-Lobato, Jos\'{e} Miguel},
  year         = {2020},
  booktitle    = {Advances in Neural Information Processing Systems},
  publisher    = {Curran Associates, Inc.},
  volume       = {33},
  pages        = {6852--6866},
}

@misc{enamine,
  title        = {Building Blocks Catalog},
  author       = {Enamine},
  year         = {2023},
  url          = {https://enamine.net/building-blocks/building-blocks-catalog},
}

@misc{dds10,
  title        = {Diversity Libraries},
  author       = {Enamine},
  year         = {2025},
  url          = {https://enamine.net/compound-libraries/diversity-libraries},
}

@inproceedings{f-rag,
  title        = {Molecule Generation with Fragment Retrieval Augmentation},
  author       = {Seul Lee and Karsten Kreis and Srimukh Prasad Veccham and Meng Liu and Danny Reidenbach and Saee Gopal Paliwal and Arash Vahdat and Weili Nie},
  year         = {2024},
  booktitle    = {The Thirty-eighth Annual Conference on Neural Information Processing Systems},
}

@article{gao2020synthesizability,
  title        = {The Synthesizability of Molecules Proposed by Generative Models},
  author       = {Gao, Wenhao and Coley, Connor W.},
  year         = {2020},
  journal      = {Journal of Chemical Information and Modeling},
  volume       = {60},
  number       = {12},
  pages        = {5714--5723},
  doi          = {10.1021/acs.jcim.0c00174},
}

@article{gauche,
  title        = {{GAUCHE}: A library for {Gaussian} processes in chemistry},
  author       = {Griffiths, Ryan-Rhys and Klarner, Leo and Moss, Henry and Ravuri, Aditya and Truong, Sang and Du, Yuanqi and Stanton, Samuel and Tom, Gary and Rankovic, Bojana and Jamasb, Arian and others},
  year         = {2024},
  journal      = {Advances in Neural Information Processing Systems},
  volume       = {36},
}

@inproceedings{gegl,
 author = {Ahn, Sungsoo and Kim, Junsu and Lee, Hankook and Shin, Jinwoo},
 booktitle = {Advances in Neural Information Processing Systems},
 pages = {12008--12021},
 publisher = {Curran Associates, Inc.},
 title = {Guiding Deep Molecular Optimization with Genetic Exploration},
 volume = {33},
 year = {2020}
}

@inproceedings{geneticgfn,
  title        = {Genetic-guided {GF}lowNets for Sample Efficient Molecular Optimization},
  author       = {Hyeonah Kim and Minsu Kim and Sanghyeok Choi and Jinkyoo Park},
  year         = {2024},
  booktitle    = {The Thirty-eighth Annual Conference on Neural Information Processing Systems},
}

@article{gobbi1998,
  title        = {Genetic optimization of combinatorial libraries},
  author       = {Gobbi, Alberto and Poppinger, Dieter},
  year         = {1998},
  journal      = {Biotechnology and Bioengineering},
  volume       = {61},
  number       = {1},
  pages        = {47--54},
  doi          = {10.1002/(SICI)1097-0290(199824)61:1<47::AID-BIT9>3.0.CO;2-Z},
}

@inproceedings{gpbo,
  title        = {Diagnosing and fixing common problems in Bayesian optimization for molecule design},
  author       = {Austin Tripp and Jos{\'e} Miguel Hern{\'a}ndez-Lobato},
  year         = {2024},
  booktitle    = {ICML 2024 AI for Science Workshop},
}

@inproceedings{gpytorch,
  title        = {{GPyTorch}: Blackbox Matrix-Matrix Gaussian Process Inference with {GPU} Acceleration},
  author       = {Gardner, Jacob R and Pleiss, Geoff and Bindel, David and Weinberger, Kilian Q and Wilson, Andrew Gordon},
  year         = {2018},
  booktitle    = {Advances in Neural Information Processing Systems},
}

@article{graphga,
  title        = {A graph-based genetic algorithm and generative model/Monte Carlo tree search for the exploration of chemical space},
  author       = {Jensen, Jan H.},
  year         = {2019},
  journal      = {Chem. Sci.},
  publisher    = {The Royal Society of Chemistry},
  volume       = {10},
  pages        = {3567--3572},
  doi          = {10.1039/C8SC05372C},
  issue        = {12},
}

@article{guacamol,
  title        = {{GuacaMol}: Benchmarking Models for de Novo Molecular Design},
  author       = {Brown, Nathan and Fiscato, Marco and Segler, Marwin H.S. and Vaucher, Alain C.},
  year         = {2019},
  journal      = {Journal of Chemical Information and Modeling},
  volume       = {59},
  number       = {3},
  pages        = {1096--1108},
  doi          = {10.1021/acs.jcim.8b00839},
}

@article{hartenfeller2011,
  title        = {A Collection of Robust Organic Synthesis Reactions for In Silico Molecule Design},
  author       = {Hartenfeller, Markus and Eberle, Martin and Meier, Peter and Nieto-Oberhuber, Cristina and Altmann, Karl-Heinz and Schneider, Gisbert and Jacoby, Edgar and Renner, Steffen},
  year         = {2011},
  journal      = {Journal of Chemical Information and Modeling},
  volume       = {51},
  number       = {12},
  pages        = {3093--3098},
  doi          = {10.1021/ci200379p},
}

@book{holland1992adaptation,
  title        = {Adaptation in natural and artificial systems: an introductory analysis with applications to biology, control, and artificial intelligence},
  author       = {Holland, John H},
  year         = {1992},
  publisher    = {MIT press},
}

@article{janus,
  title        = {Parallel tempered genetic algorithm guided by deep neural networks for inverse molecular design},
  author       = {Nigam, AkshatKumar and Pollice, Robert and Aspuru-Guzik, Alán},
  year         = {2022},
  journal      = {Digital Discovery},
  publisher    = {RSC},
  volume       = {1},
  pages        = {390--404},
  doi          = {10.1039/D2DD00003B},
  issue        = {4},
}

@article{kim2021polymer,
  title        = {Polymer design using genetic algorithm and machine learning},
  author       = {Chiho Kim and Rohit Batra and Lihua Chen and Huan Tran and Rampi Ramprasad},
  year         = {2021},
  journal      = {Computational Materials Science},
  volume       = {186},
  pages        = {110067},
  doi          = {10.1016/j.commatsci.2020.110067},
}

@article{kneilding2024augmenting,
  title        = {Augmenting genetic algorithms with machine learning for inverse molecular design},
  author       = {Kneiding, Hannes and Balcells, David},
  year         = {2024},
  journal      = {Chem. Sci.},
  publisher    = {The Royal Society of Chemistry},
  volume       = {15},
  pages        = {15522--15539},
  doi          = {10.1039/D4SC02934H},
  issue        = {38},
}

@article{levin2023computer,
  title        = {Computer-aided evaluation and exploration of chemical spaces constrained by reaction pathways},
  author       = {Levin, Itai and Fortunato, Michael E. and Tan, Kian L. and Coley, Connor W.},
  year         = {2023},
  journal      = {AIChE Journal},
  volume       = {69},
  number       = {12},
  pages        = {e18234},
  doi          = {10.1002/aic.18234},
}

@article{li2018multi,
  title        = {Multi-objective de novo drug design with conditional graph generative model},
  author       = {Li, Yibo and Zhang, Liangren and Liu, Zhenming},
  year         = {2018},
  month        = {Jul},
  day          = {24},
  journal      = {Journal of Cheminformatics},
  volume       = {10},
  number       = {1},
  pages        = {33},
  doi          = {10.1186/s13321-018-0287-6},
}

@article{litpcba,
  title        = {{LIT-PCBA}: An Unbiased Data Set for Machine Learning and Virtual Screening},
  author       = {Tran-Nguyen, Viet-Khoa and Jacquemard, C{\'e}lien and Rognan, Didier},
  year         = {2020},
  journal      = {Journal of Chemical Information and Modeling},
  volume       = {60},
  number       = {9},
  pages        = {4263--4273},
  doi          = {10.1021/acs.jcim.0c00155},
}

@inproceedings{llmsynplanner,
  title        = {{LLM}-Augmented Chemical Synthesis and Design Decision Programs},
  author       = {Haorui Wang and Jeff Guo and Lingkai Kong and Rampi Ramprasad and Philippe Schwaller and Yuanqi Du and Chao Zhang},
  year         = {2025},
  booktitle    = {Towards Agentic AI for Science: Hypothesis Generation, Comprehension, Quantification, and Validation},
}

@inproceedings{moleculechef,
  title        = {A Model to Search for Synthesizable Molecules},
  author       = {Bradshaw, John and Paige, Brooks and Kusner, Matt J and Segler, Marwin and Hern\'{a}ndez-Lobato, Jos\'{e} Miguel},
  year         = {2019},
  booktitle    = {Advances in Neural Information Processing Systems},
  publisher    = {Curran Associates, Inc.},
  volume       = {32},
  pages        = {},
}

@article{molga,
  title        = {Genetic algorithms are strong baselines for molecule generation},
  author       = {Austin Tripp and José Miguel Hernández-Lobato},
  year         = {2023},
  eprint       = {2310.09267},
  archiveprefix = {arXiv},
  primaryclass = {cs.NE},
  journal     = {arXiv preprint arXiv:2310.09267}
}

@inproceedings{nam,
  title        = {Neural Additive Models: Interpretable Machine Learning with Neural Nets},
  author       = {Rishabh Agarwal and Levi Melnick and Nicholas Frosst and Xuezhou Zhang and Ben Lengerich and Rich Caruana and Geoffrey Hinton},
  year         = {2021},
  booktitle    = {Advances in Neural Information Processing Systems},
}

@inproceedings{
neuralga,
title={Neural Genetic Search in Discrete Spaces},
author={Hyeonah Kim and Sanghyeok Choi and Jiwoo Son and Jinkyoo Park and Changhyun Kwon},
booktitle={Forty-second International Conference on Machine Learning},
year={2025},
}

@inproceedings{nigam2020augmenting,
  title        = {Augmenting Genetic Algorithms with Deep Neural Networks for Exploring the Chemical Space},
  author       = {AkshatKumar Nigam and Pascal Friederich and Mario Krenn and Alan Aspuru-Guzik},
  year         = {2020},
  booktitle    = {International Conference on Learning Representations},
}

@article{nigam2024emitters,
  title        = {Artificial design of organic emitters via a genetic algorithm enhanced by a deep neural network},
  author       = {Nigam, AkshatKumar and Pollice, Robert and Friederich, Pascal and Aspuru-Guzik, Alán},
  year         = {2024},
  journal      = {Chem. Sci.},
  publisher    = {The Royal Society of Chemistry},
  volume       = {15},
  pages        = {2618--2639},
  doi          = {10.1039/D3SC05306G},
  issue        = {7},
}

@article{nsgaii,
  title        = {A fast and elitist multiobjective genetic algorithm: {NSGA-II}},
  author       = {Deb, K. and Pratap, A. and Agarwal, S. and Meyarivan, T.},
  year         = {2002},
  journal      = {IEEE Transactions on Evolutionary Computation},
  volume       = {6},
  number       = {2},
  pages        = {182--197},
  doi          = {10.1109/4235.996017},
}

@inproceedings{pgfs,
  title        = {Learning to Navigate The Synthetically Accessible Chemical Space Using Reinforcement Learning},
  author       = {Gottipati, Sai Krishna and Sattarov, Boris and Niu, Sufeng and Pathak, Yashaswi and Wei, Haoran and Liu, Shengchao and Liu, Shengchao and Blackburn, Simon and Thomas, Karam and Coley, Connor and Tang, Jian and Chandar, Sarath and Bengio, Yoshua},
  year         = {2020},
  month        = {13--18 Jul},
  booktitle    = {Proceedings of the 37th International Conference on Machine Learning},
  publisher    = {PMLR},
  series       = {Proceedings of Machine Learning Research},
  volume       = {119},
  pages        = {3668--3679},
}

@inproceedings{pmo,
  title        = {Sample Efficiency Matters: A Benchmark for Practical Molecular Optimization},
  author       = {Gao, Wenhao and Fu, Tianfan and Sun, Jimeng and Coley, Connor},
  year         = {2022},
  booktitle    = {Advances in Neural Information Processing Systems},
  publisher    = {Curran Associates, Inc.},
  volume       = {35},
  pages        = {21342--21357},
}

@inproceedings{pocket2mol,
  title        = {{Pocket2Mol}: Efficient Molecular Sampling Based on 3D Protein Pockets},
  author       = {Xingang Peng and Shitong Luo and Jiaqi Guan and Qi Xie and Jian Peng and Jianzhu Ma},
  year         = {2022},
  booktitle    = {International Conference on Machine Learning},
}

@article{qed,
  title        = {Quantifying the chemical beauty of drugs},
  author       = {Bickerton, G. Richard and Paolini, Gaia V. and Besnard, J{\'e}r{\'e}my and Muresan, Sorel and Hopkins, Andrew L.},
  year         = {2012},
  month        = {Feb},
  day          = {01},
  journal      = {Nature Chemistry},
  volume       = {4},
  number       = {2},
  pages        = {90--98},
  doi          = {10.1038/nchem.1243},
}

@article{quickvina2,
  title        = {Fast, accurate, and reliable molecular docking with {QuickVina} 2},
  author       = {Alhossary, Amr and Handoko, Stephanus Daniel and Mu, Yuguang and Kwoh, Chee-Keong},
  year         = {2015},
  month        = {02},
  journal      = {Bioinformatics},
  volume       = {31},
  number       = {13},
  pages        = {2214--2216},
  doi          = {10.1093/bioinformatics/btv082},
}

@inproceedings{ranknet,
  title        = {Learning to rank using gradient descent},
  author       = {Burges, Chris and Shaked, Tal and Renshaw, Erin and Lazier, Ari and Deeds, Matt and Hamilton, Nicole and Hullender, Greg},
  year         = {2005},
  booktitle    = {Proceedings of the 22nd International Conference on Machine Learning},
  publisher    = {Association for Computing Machinery},
  address      = {New York, NY, USA},
  pages        = {89–96},
  doi          = {10.1145/1102351.1102363},
  numpages     = {8},
}

@misc{rdkit,
  title        = {{RDKit}: Open-source cheminformatics},
  author       = {Landrum, Greg and others},
  year         = {2006},
}

@article{reactor,
  title        = {Molecular Design in Synthetically Accessible Chemical Space via Deep Reinforcement Learning},
  author       = {Horwood, Julien and Noutahi, Emmanuel},
  year         = {2020},
  month        = {Dec},
  day          = {29},
  journal      = {ACS Omega},
  publisher    = {American Chemical Society},
  volume       = {5},
  number       = {51},
  pages        = {32984--32994},
  doi          = {10.1021/acsomega.0c04153},
}

@article{recap,
  title        = {{RECAP}- Retrosynthetic Combinatorial Analysis Procedure: A Powerful New Technique for Identifying Privileged Molecular Fragments with Useful Applications in Combinatorial Chemistry},
  author       = {Lewell, Xiao Qing and Judd, Duncan B. and Watson, Stephen P. and Hann, Michael M.},
  year         = {1998},
  journal      = {Journal of Chemical Information and Computer Sciences},
  volume       = {38},
  number       = {3},
  pages        = {511--522},
  doi          = {10.1021/ci970429i},
}

@article{reinvent,
  title        = {Molecular de-novo design through deep reinforcement learning},
  author       = {Olivecrona, Marcus and Blaschke, Thomas and Engkvist, Ola and Chen, Hongming},
  year         = {2017},
  month        = {Sep},
  day          = {04},
  journal      = {Journal of Cheminformatics},
  volume       = {9},
  number       = {1},
  pages        = {48},
  doi          = {10.1186/s13321-017-0235-x},
}

@inproceedings{rgfn,
  title        = {{RGFN}: Synthesizable Molecular Generation Using GFlowNets},
  author       = {Koziarski, Micha{\l}  and Rekesh, Andrei and Shevchuk, Dmytro and van der Sloot, Almer and Gai\'{n}ski, Piotr and Bengio, Yoshua and Liu, Cheng-Hao and Tyers, Mike and Batey, Robert A.},
  year         = {2024},
  booktitle    = {Advances in Neural Information Processing Systems},
  publisher    = {Curran Associates, Inc.},
  volume       = {37},
  pages        = {46908--46955},
}

@inproceedings{rga,
  title        = {Reinforced Genetic Algorithm for Structure-based Drug Design},
  author       = {Fu, Tianfan and Gao, Wenhao and Coley, Connor and Sun, Jimeng},
  year         = {2022},
  booktitle    = {Advances in Neural Information Processing Systems},
  publisher    = {Curran Associates, Inc.},
  volume       = {35},
  pages        = {12325--12338},
}

@inproceedings{robinson2021contrastive,
  title        = {Contrastive Learning with Hard Negative Samples},
  author       = {Joshua David Robinson and Ching-Yao Chuang and Suvrit Sra and Stefanie Jegelka},
  year         = {2021},
  booktitle    = {International Conference on Learning Representations},
}

@inproceedings{rxnflow,
  title        = {Generative Flows on Synthetic Pathway for Drug Design},
  author       = {Seonghwan Seo and Minsu Kim and Tony Shen and Martin Ester and Jinkyoo Park and Sungsoo Ahn and Woo Youn Kim},
  year         = {2025},
  booktitle    = {The Thirteenth International Conference on Learning Representations},
}

@article{sascore,
  title        = {Estimation of synthetic accessibility score of drug-like molecules based on molecular complexity and fragment contributions},
  author       = {Ertl, Peter and Schuffenhauer, Ansgar},
  year         = {2009},
  month        = {Jun},
  day          = {10},
  journal      = {Journal of Cheminformatics},
  volume       = {1},
  number       = {1},
  pages        = {8},
  doi          = {10.1186/1758-2946-1-8},
}

@article{saturn,
  title        = {Saturn: Sample-efficient Generative Molecular Design using Memory Manipulation},
  author       = {Jeff Guo and Philippe Schwaller},
  year         = {2024},
  eprint       = {2405.17066},
  archiveprefix = {arXiv},
  primaryclass = {q-bio.BM},
  journal = {arXiv preprint arXiv:2405.17066}
}

@article{scaffold,
  title        = {The Properties of Known Drugs. 1. Molecular Frameworks},
  author       = {Bemis, Guy W. and Murcko, Mark A.},
  year         = {1996},
  journal      = {Journal of Medicinal Chemistry},
  volume       = {39},
  number       = {15},
  pages        = {2887--2893},
  doi          = {10.1021/jm9602928},
}

@article{selfies1,
  title        = {Self-referencing embedded strings ({SELFIES}): A 100% robust molecular string representation},
  author       = {Krenn, Mario and Häse, Florian and Nigam, AkshatKumar and Friederich, Pascal and Aspuru-Guzik, Alan},
  year         = {2020},
  month        = {oct},
  journal      = {Machine Learning: Science and Technology},
  publisher    = {IOP Publishing},
  volume       = {1},
  number       = {4},
  pages        = {045024},
  doi          = {10.1088/2632-2153/aba947},
}

@article{selfies2,
  title        = {Recent advances in the self-referencing embedded strings ({SELFIES}) library},
  author       = {Lo, Alston and Pollice, Robert and Nigam, AkshatKumar and White, Andrew D. and Krenn, Mario and Aspuru-Guzik, Alán},
  year         = {2023},
  journal      = {Digital Discovery},
  publisher    = {RSC},
  volume       = {2},
  pages        = {897--908},
  doi          = {10.1039/D3DD00044C},
  issue        = {4},
}

@article{seumer2024beyond,
  title        = {Beyond Predefined Ligand Libraries: A Genetic Algorithm Approach for De Novo Discovery of Catalysts for the Suzuki Coupling Reactions},
  author       = {Seumer, Julius and Jensen, Jan H.},
  year         = {2024},
  journal      = {ChemRxiv},
  doi          = {10.26434/chemrxiv-2024-9xh38},
}

@article{smilesDrawer,
  title        = {{SmilesDrawer}: Parsing and Drawing SMILES-Encoded Molecular Structures Using Client-Side JavaScript},
  author       = {Probst, Daniel and Reymond, Jean-Louis},
  year         = {2018},
  journal      = {Journal of Chemical Information and Modeling},
  volume       = {58},
  number       = {1},
  pages        = {1--7},
  doi          = {10.1021/acs.jcim.7b00425},
}

@article{smileslstm,
  title        = {Generating Focused Molecule Libraries for Drug Discovery with Recurrent Neural Networks},
  author       = {Segler, Marwin H. S. and Kogej, Thierry and Tyrchan, Christian and Waller, Mark P.},
  year         = {2018},
  journal      = {ACS Central Science},
  volume       = {4},
  number       = {1},
  pages        = {120--131},
  doi          = {10.1021/acscentsci.7b00512},
}

@article{stanley2023fake,
  title        = {Fake it until you make it? Generative de novo design and virtual screening of synthesizable molecules},
  author       = {Megan Stanley and Marwin Segler},
  year         = {2023},
  journal      = {Current Opinion in Structural Biology},
  volume       = {82},
  pages        = {102658},
  doi          = {10.1016/j.sbi.2023.102658},
}

@inproceedings{synflownet,
  title        = {{SynFlowNet}: Design of Diverse and Novel Molecules with Synthesis Constraints},
  author       = {Miruna Cretu and Charles Harris and Ilia Igashov and Arne Schneuing and Marwin Segler and Bruno Correia and Julien Roy and Emmanuel Bengio and Pietro Lio},
  year         = {2025},
  booktitle    = {The Thirteenth International Conference on Learning Representations},
}

@article{synformer,
  title        = {Generative Artificial Intelligence for Navigating Synthesizable Chemical Space},
  author       = {Wenhao Gao and Shitong Luo and Connor W. Coley},
  year         = {2024},
  eprint       = {2410.03494},
  archiveprefix = {arXiv},
  primaryclass = {cs.LG},
  journal = {arXiv preprint arXiv:2410.03494}
}

@article{synllama,
  title        = {{SynLlama}: Generating Synthesizable Molecules and Their Analogs with Large Language Models},
  author       = {Kunyang Sun and Dorian Bagni and Joseph M. Cavanagh and Yingze Wang and Jacob M. Sawyer and Andrew Gritsevskiy and Teresa Head-Gordon},
  year         = {2025},
  eprint       = {2503.12602},
  archiveprefix = {arXiv},
  primaryclass = {cs.LG},
  journal = {arXiv preprint arXiv:2503.12602}
}

@inproceedings{synnet,
  title        = {Amortized Tree Generation for Bottom-up Synthesis Planning and Synthesizable Molecular Design},
  author       = {Wenhao Gao and Roc{\'\i}o Mercado and Connor W. Coley},
  year         = {2022},
  booktitle    = {International Conference on Learning Representations},
}

@article{synopsis,
  title        = {{SYNOPSIS}: SYNthesize and OPtimize System in Silico},
  author       = {Vinkers, H. Maarten and de Jonge, Marc R. and Daeyaert, Frederik F. D. and Heeres, Jan and Koymans, Lucien M. H. and van Lenthe, Joop H. and Lewi, Paul J. and Timmerman, Henk and Van Aken, Koen and Janssen, Paul A. J.},
  year         = {2003},
  journal      = {Journal of Medicinal Chemistry},
  volume       = {46},
  number       = {13},
  pages        = {2765--2773},
  doi          = {10.1021/jm030809x},
}

@article{synthemol,
  title        = {Generative {AI} for designing and validating easily synthesizable and structurally novel antibiotics},
  author       = {Swanson, Kyle and Liu, Gary and Catacutan, Denise B. and Arnold, Autumn and Zou, James and Stokes, Jonathan M.},
  year         = {2024},
  month        = {Mar},
  day          = {01},
  journal      = {Nature Machine Intelligence},
  volume       = {6},
  number       = {3},
  pages        = {338--353},
  doi          = {10.1038/s42256-024-00809-7},
}

@inproceedings{synthesisnet,
  title        = {Procedural Synthesis of Synthesizable Molecules},
  author       = {Michael Sun and Alston Lo and Minghao Guo and Jie Chen and Connor W. Coley and Wojciech Matusik},
  year         = {2025},
  booktitle    = {The Thirteenth International Conference on Learning Representations},
}

@article{tango,
  title        = {Directly optimizing for synthesizability in generative molecular design using retrosynthesis models},
  author       = {Guo, Jeff and Schwaller, Philippe},
  year         = {2025},
  journal      = {Chem. Sci.},
  publisher    = {The Royal Society of Chemistry},
  volume       = {16},
  pages        = {6943--6956},
  doi          = {10.1039/D5SC01476J},
  issue        = {16},
}

@article{tdc,
  title        = {{Therapeutics Data Commons}: Machine Learning Datasets and Tasks for Drug Discovery and Development},
  author       = {Huang, Kexin and Fu, Tianfan and Gao, Wenhao and Zhao, Yue and Roohani, Yusuf and Leskovec, Jure and Coley, Connor W and Xiao, Cao and Sun, Jimeng and Zitnik, Marinka},
  year         = {2021},
  journal      = {Proceedings of Neural Information Processing Systems, NeurIPS Datasets and Benchmarks},
}

@article{terfloth2001neural,
  title        = {Neural networks and genetic algorithms in drug design},
  author       = {Terfloth, Lothar and Gasteiger, Johann},
  year         = {2001},
  journal      = {Drug Discovery Today},
  publisher    = {Elsevier},
  volume       = {6},
  pages        = {102--108},
}

@inproceedings{tripp2021fresh,
  title        = {A Fresh Look at De Novo Molecular Design Benchmarks},
  author       = {Austin Tripp and Gregor N. C. Simm and Jos{\'e} Miguel Hern{\'a}ndez-Lobato},
  year         = {2021},
  booktitle    = {NeurIPS 2021 AI for Science Workshop},
}

@article{unidock,
  title        = {{Uni-Dock}: {GPU}-Accelerated Docking Enables Ultralarge Virtual Screening},
  author       = {Yu, Yuejiang and Cai, Chun and Wang, Jiayue and Bo, Zonghua and Zhu, Zhengdan and Zheng, Hang},
  year         = {2023},
  month        = {Jun},
  day          = {13},
  journal      = {Journal of Chemical Theory and Computation},
  publisher    = {American Chemical Society},
  volume       = {19},
  number       = {11},
  pages        = {3336--3345},
  doi          = {10.1021/acs.jctc.2c01145},
}

@article{zinc,
  title        = {{ZINC20}—A Free Ultralarge-Scale Chemical Database for Ligand Discovery},
  author       = {Irwin, John J. and Tang, Khanh G. and Young, Jennifer and Dandarchuluun, Chinzorig and Wong, Benjamin R. and Khurelbaatar, Munkhzul and Moroz, Yurii S. and Mayfield, John and Sayle, Roger A.},
  year         = {2020},
  journal      = {Journal of Chemical Information and Modeling},
  volume       = {60},
  number       = {12},
  pages        = {6065--6073},
  doi          = {10.1021/acs.jcim.0c00675},
}

@inproceedings{tartarus,
 author = {Nigam, AkshatKumar and Pollice, Robert and Tom, Gary and Jorner, Kjell and Willes, John and Thiede, Luca and Kundaje, Anshul and Aspuru-Guzik, Alan},
 booktitle = {Advances in Neural Information Processing Systems},
 pages = {3263--3306},
 publisher = {Curran Associates, Inc.},
 title = {Tartarus: A Benchmarking Platform for Realistic And Practical Inverse Molecular Design},
 volume = {36},
 year = {2023}
}
